\documentclass[twoside]{article}

\usepackage[accepted]{aistats2020}
\usepackage{hyperref}
\usepackage{url}
\usepackage{amsmath}
\usepackage{amssymb}
\usepackage{amsthm}
\usepackage{caption}
\usepackage{subcaption}
\usepackage{graphicx}
\usepackage[dvipsnames]{xcolor}
\usepackage{xfrac}
\usepackage{tabularx}
\usepackage{booktabs} 
\usepackage{hyperref}
\usepackage[round]{natbib}
\usepackage{dblfloatfix}

\newcommand\T{\rule{0pt}{2.6ex}}       

\newtheorem{theorem}{Theorem}[section]

\newtheorem{lemma}[theorem]{Lemma}
\theoremstyle{definition}

\newtheorem{innercustomthm}{Theorem}
\newenvironment{customthm}[1]
  {\renewcommand\theinnercustomthm{#1}\innercustomthm}
  {\endinnercustomthm}

\newtheorem{innercustomlem}{Lemma}
\newenvironment{customlem}[1]
  {\renewcommand\theinnercustomlem{#1}\innercustomlem}
  {\endinnercustomlem}

\newcommand{\bw}{\boldsymbol{w}}
\newcommand{\bu}{\boldsymbol{u}}
\newcommand{\bx}{\boldsymbol{x}}
\newcommand{\bW}{\boldsymbol{W}}
\newcommand{\bU}{\boldsymbol{U}}
\newcommand{\bX}{\boldsymbol{X}}

\newcommand{\cW}{\mathcal{W}}
\newcommand{\cC}{\mathcal{C}}
\newcommand{\cO}{\mathcal{O}}
\newcommand{\cD}{\mathcal{D}}
\newcommand{\cF}{\mathcal{F}}
\newcommand{\cL}{\mathcal{L}}

\begin{document}

%

%

\twocolumn[

\aistatstitle{The role of invariance in spectral complexity-based generalization bounds}

\aistatsauthor{$\mathrm{\textbf{Konstantinos Pitas}}^1$, $\mathrm{\textbf{Andreas Loukas}}^1$, $\mathrm{\textbf{Mike Davies}}^2$, $\mathrm{\textbf{Pierre Vandergheynst}}^1$}

\aistatsaddress{${}^1\mathrm{EPFL}$,${}^2\mathrm{University\; of\; Edinburgh}$}]

\begin{abstract}
Deep convolutional neural networks (CNNs) have been shown to be able to fit a random labeling over data while still being able to generalize well for normal labels. Describing CNN capacity through a posteriory measures of complexity has been recently proposed to tackle this apparent paradox. These complexity measures are usually validated by showing that they correlate empirically with GE; being empirically larger for networks trained on random vs normal labels. Focusing on the case of \emph{spectral} complexity we investigate theoretically and empirically the insensitivity of the complexity measure to invariances relevant to CNNs, and show several limitations of spectral complexity that occur as a result. For a specific formulation of spectral complexity we show that it results in the same upper bound complexity estimates for convolutional and locally connected architectures (which don't have the same favorable invariance properties). This is contrary to common intuition and empirical results.
\end{abstract}

\section{Introduction}
\label{sec:intro}

Standard deep convolutional networks (CNNs) have the capacity to fit a random labelling over data~\citep{zhang2016understanding}. At the same time, these same networks generalize well for real labels. 
In fact, any measure of model complexity which is uniform across all functions representable by a given architecture is doomed to provide contradictory measurements \citep{bartlett2017spectrally,arora2018stronger,neyshabur2015norm}. A good measure of complexity should allow for high complexity models for difficult datasets (random labels) and low complexity models for easier datasets (real labels).

\begin{figure}[t!]
\hspace{-7.5mm}
\includegraphics[width=1.19\columnwidth]{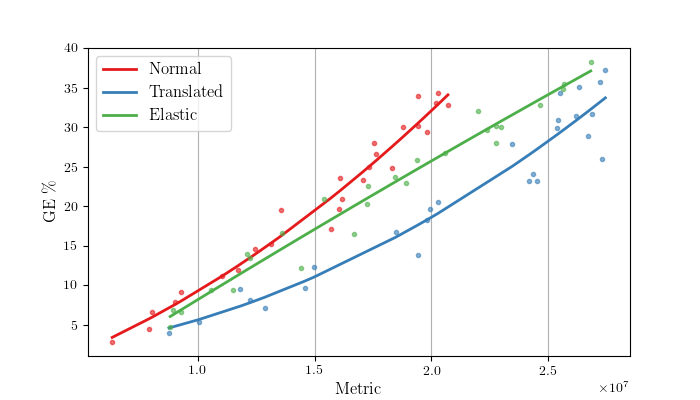}
\caption{
Spectral complexity analyses of generalization error are insensitive to the known invariances of CNNs.
We use 10$^4$ training and 10$^4$ testing images of Cifar-10 as our \textit{Control} dataset and create two additional datasets called \textit{Elastic Cifar} and \textit{Translated Cifar} by removing half of the training and testing sets and replacing them, respectively, with random elastic deformations and translations of the remaining images. 
We observe that, for constant spectral complexity (gray vertical lines), trained networks exhibit different generalization error for different datasets.\vspace{0mm}}
\label{fig:conv_complete}
\end{figure}

Inspired by this realization, researchers interested in generalization error (GE) bounds have recently focused on \emph{spectral complexity} \citep{bartlett2017spectrally} normalized by the {margin}. Spectral complexity consists usually of the product of the spectral or other norms of the different layer weight matrices. The average margin quantifies the confidence of the classifier: it is the average difference between the first and second most probable class estimates per sample. Different techniques for bounding the sample complexity include robustness, PAC-Bayes and Rademacher complexity \citep{sokolic2016robust,bartlett2017spectrally,neyshabur2017pac,neyshabur2015norm,golowich2017size}. While derived for general weight matrices, these bounds are often applied to deep convolutional networks. In this context spectral complexity 
has been shown to correlate empirically with the generalization error in a number of works \citep{neyshabur2017exploring,bartlett2017spectrally}. For the same network, this measure of model complexity has high values when the network is trained on data with random labels and considerably lower values for real labels. A number of other measures \citep{wang2018identifying,keskar2016large,thomas2019information,wei2019data,jiang2018predicting,liang2019fisher,arora2018stronger} have been proposed, with some correlating better than others with generalization error. 

On a more fundamental level simple correlation with generalization error is unsatisfying given that deep neural networks are increasingly being deployed in critical environments such as healthcare, finance and policing where they can potentially make life altering decisions. Spectral complexity-based bounds in particular have demonstrated empirically by \cite{arora2018stronger} to be vacuous by several orders of magnitude. Consequently, some works aimed to obtain non-vacuous bounds through optimisation of a stochastic DNN \cite{dziugaite2017computing} or compression of a DNN \citep{zhou2018non}. The main issue with such analyses is that a one to one correspondence cannot be established between the optimised or compressed architecture and the original one. Thus, besides not being sufficiently tight, the derived bounds do not apply to the original classifier. 

\textbf{The role of invariance.} 
Invariances are widely considered to be crucial in DNN design \citep{bengio2013representation}. On the theoretical side, some CNNs have been proven to be invariant to translations and stable to deformations~\citep{mallat2016understanding,wiatowski2018mathematical}. Also, CNNs, after training, empirically appear to be invariant to much more complex transformations on the data, such as adding sunglasses to faces \citep{radford2015unsupervised2}.

Interestingly, while it is generally agreed that invariance to symmetries in the image data is a key property of modern deep convolutional neural networks, the role of invariances is conspicuously absent from the generalization literature. \cite{achille2018emergence} showed that low information content in the network weights corresponds to learning invariant signal representations to various nuisance latent parameters. Their work however does not result in a meaningful generalization bound. Further, \cite{sokolic2016generalization} demonstrated that classifiers that are invariant (to a set of discrete transformations of input signals) can potentially have a much lower GE than non-invariant ones. 

Similarly, due to the non-trivial correlations between filters, the generalization capacity of deep CNNs has been rarely studied. Works such as \cite{zhou2018understanding}, \cite{du2017gradient},\cite{arora2018stronger},\cite{long2019size},\cite{li2018tighter} are typically very involved, analyze greatly restricted settings and do not seem to lead to non-vacuous generalization bounds or to any new intuition apart from better parameter counting. Crucially an open question remains:

\textit{To what extent do existing bounds and complexity measures incorporate the invariance properties induced by deep convolutional architectures?}


\textbf{Contributions.} Focusing on the popular case of spectral complexity:

\begin{itemize}
\item We confirm empirically that spectral complexity bounds fail to capture the invariance properties of CNNs to data symmetries, such as elastic deformations and translations. As seen in Figure \ref{fig:conv_complete}, CNNs with the same spectral complexity exhibit different GE when we augment the dataset with perturbations to which the convolutional architecture is inherently invariant. Our experiments suggest that these conclusions are not unique to our approach, but apply to spectral complexity-based generalization bounds in general~\citep{bartlett2002rademacher,sokolic2016robust,bartlett2017spectrally,neyshabur2017pac,neyshabur2015norm,golowich2017size}. We conclude that more research should be conducted in incorporating invariance properties in GE analyses.

\item We analyze the case of \emph{locally-connected} layers, i.e., layers constructed to have the same support as convolutional layers but which don't employ weight sharing. As such deep locally connected networks should not have the desired invariance properties of stacked convolutions. Counter-intuitively, we arrive to the same generalization error guarantees as convolutional architectures (up to negligible factors that are artifacts of the derivation). Our experiments indicate that crucial quantities in the bound are tight, pointing to an inherent shortcoming of spectral complexity.
\end{itemize}

While we empirically test only certain spectral complexity based bounds, our results should be meaningful for most current bounds. These typically hold for \emph{any} data generating distribution, and therefore should ignore the input data structure and the corresponding invariance properties of modern CNNs.


\section{Spectral complexity metrics}

Let $\cD$ be a distribution over samples $\bx$ and labels $y$. We consider the standard classification problem in which a $k$-class classifier $f_{\bw}: \mathbb{R}^n \rightarrow \mathbb{R}^k$ parameterized by $\bw$ is used to map input vectors $\bx$ to a $k$-dimensional vector, encoding class membership. 

We may encode the confidence of the classifier by incorporating a dependence on a desired margin $\gamma>0$. Then, the \emph{$\gamma$-margin classification loss} is defined as 
%
%
\begin{align*}
    L_{\gamma}(f_{\bw}) := \mathbb{P}_{(\bx,y)\sim \cD} \left[ f_{\bw}(\bx)[y] \leq \gamma + \max_{j \neq y} f_{\bw}(\bx)[j] \right].
\end{align*} 
Note that we easily recover the standard classification loss definition $L_0(f_{\bw})$ by setting $\gamma = 0$. 
Our objective is to obtain bounds of the generalization error GE:
\begin{align}
    L_{0}(f_{\bw}) \leq \hat{L}_{\gamma}(f_{\bw}) + \text{GE},
\end{align}
where 
$$
\hat{L}_{\gamma}(f_{\bw}) = \sum_{i = 1}^m \frac{ 1 \{f_{\bw}(\bx_i)[y_i]< \max_{j \neq y_i} f_{\bw}(\bx_i)[j] \} }{m}
$$
is the \emph{empirical} loss computed over a random training set of size $m$. For easy reference, we summarize some of the most crucial definitions in Table~\ref{tab:notation}.

\begin{table}[t]
\begin{center}
\begin{small}
\begin{tabular}{ l l   }
  \toprule
  symbol & meaning \\ 
  \midrule
  $n$ & input image size is $n=n_1\times n_2$ \\   
  $f_{\bw}$ & neural network parameterized by $\bw$ \\
  $k$ & classes \\   
  $m$ & training set size \\   
  $d$ & network depth \\   
  $\cF$ & set of fully connected layers \\   
  $\cC$ & set of convolutional \\
  $\cL$ & set of locally-connected layers \\
  $\bW_l$ & weigh matrix of $l$-th layer \\   
  $q_l\times q_l$ & filter support in $l$-th convolutional layer  \\   
  $b_l$ & output channels in $l$-th convolutional layer\\   
\bottomrule
\end{tabular}
\end{small}
\end{center}
\vskip -0.1in
\caption{Main notation} \label{tab:notation} 
\vskip -0.1in
\end{table}

\textbf{Recent advances.} A variety of techniques, based on VC dimension, Rademacher complexity, and PAC-Bayes type arguments have been employed in the attempt to understand the generalization error of neural networks. Intriguingly, in a number of recent works the generalization error of a $d$ layer neural network with layer weights $\cW = \{\bW_1, \bW_2, \ldots, \bW_d \} $ is expressed as       
\begin{equation}
    L_0(f_{\bw}) \leq \hat{L}_{\gamma}(f_{\bw}) + \tilde{\cO}\left(\frac{ \Psi_f R_{\cW}}{ \gamma \sqrt{m}} + \Phi_f \right), 
\end{equation}
with terms $\Psi_f$ and $\Phi_f$ being {architecture}-dependent and only $R_{\cW}$ depending solely on the network weights. The latter term has been referred to as the \emph{spectral complexity} of a neural network~\citep{bartlett2017spectrally} and can be defined as 
\begin{align}
    R_{\cW} := \prod_{l=1}^d||\bW_l||_2 \, \left( \sum_{l=1}^d  \frac{||\bW_l||_F^2}{||\bW_l||_2^2}\right)^{\hspace{-1mm}\sfrac{1}{2}}.
    \label{eq:spectral_complexity}
\end{align}
To be precise, the aforementioned definition corresponds to the one derived in a PAC-Bayes framework~\citep{neyshabur2017pac} together with $\Psi_f=d\sqrt{h}$. \cite{bartlett2017spectrally} proposed the measure:
\begin{align}
    R'_{\cW} := \prod_{l=1}^d ||\bW_l||_2 \left( \sum_{l=1}^d \frac{||\bW_l^\top ||_{2,1}^{\sfrac{2}{3}}}{||\bW_l||_2^{\sfrac{2}{3}}}\right)^{\hspace{-1mm}\sfrac{3}{2}},  
    \label{eq:spectral_complexity_bartlett}
\end{align}
obtained using an involved covering argument. Here, the network weights are contrasted to some reference weights $\tilde{\bW}_l$ and the Frobenius norm is substituted by the $(2,1)$-matrix norm defined as $\|\bX\|_{2,1} = \| \|\bX_{:,1}\|_2, \ldots, \|\bX_{:,n_2}\|_2 \|_{1}$ for $\bX \in \mathbb{R}^{n_1 \times n_2}$. In the similar works of \cite{bartlett2002rademacher} and \cite{neyshabur2015norm}, the authors use the $||\cdot||_{1,\infty}$ norm and the $||\cdot||_{F}$ norm, respectively. 

Several experiments have shown that spectral complexity generally correlates empirically with the true generalization error as quantified by held out data. Furthermore, the metric is large for difficult datasets while it is small for easy datasets. Intuitively, spectral complexity is related to how robust a model is when adding noise to its layers. A simple model will be more robust to noise and can be seen as laying on a flat minimum; even moving it a way by a large quantity from the minimum center, the loss will remain approximately the same.

\section{Insensitivity of spectral complexity to data manifold symmetries}

We start by observing that spectral complexity-based measures feature a strong dependence on the $B$-stable rank of the weight matrices involved, given by
$$
 \sigma_B(\bW_{\ell}) = \frac{\|\bW_l\|_B^2 }{ \| \bW_l\|_2^2}, 
$$
where $B$ stands for a generic norm, such as the Frobenius norm in $R_{\cW}$ and the $(2,1)$ norm in $R_{\cW}'$ (see respectively \eqref{eq:spectral_complexity} and \eqref{eq:spectral_complexity_bartlett})~\citep{arora2018stronger}.
The stable rank gives a robust estimate of the degrees of freedom of a matrix: roughly, an $n_1\times n_2$ matrix $\bW$ with constant stable rank has $\cO(n_1 + n_2)$ degrees of freedom, instead of $\cO(n_1 n_2)$ as usual. 

This interpretation should give us a pause for thought: bounds based on spectral complexity (and incorporating the $B$-stable rank) appear to be sophisticated {parameter counting} techniques, able to adapt to different neural network realizations. As such, they should {in principle} not be able to capture the complex interactions between data symmetries and CNN invariance to these symmetries. 

\begin{figure*}[t!]
\centering
\hspace{-2mm}
\begin{subfigure}{.7\textwidth}
  \centering
  \includegraphics[trim=16mm 0 15mm 10mm,clip,scale=0.55]{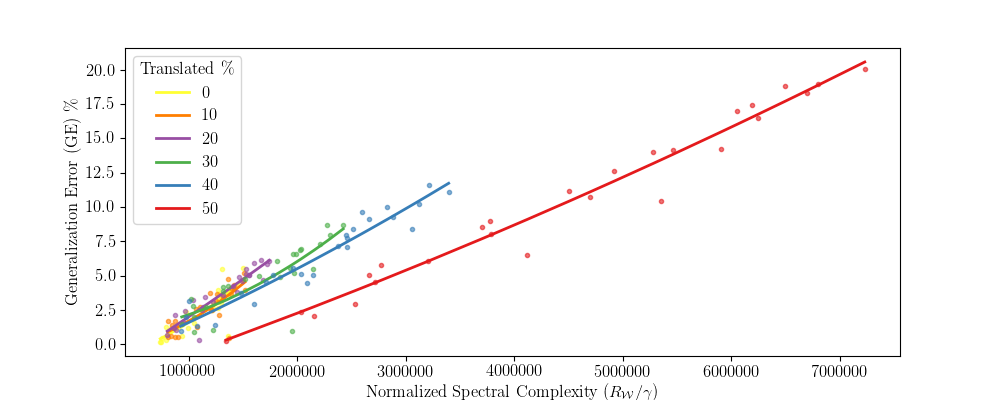}
  \caption{}
  \label{fig:conv_trans_var_metric}
\end{subfigure}%
\begin{subfigure}{.3\textwidth}
  \centering
  \includegraphics[trim=1mm 0 0mm 10mm, clip,scale=0.55]{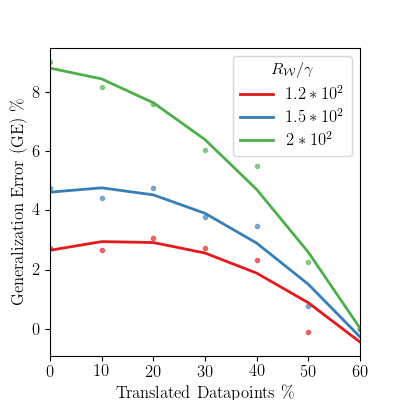}
  \caption{}
  \label{fig:conv_trans_fixed_metric}
\end{subfigure}
\\
\vspace{1mm}
\hspace{-2mm}
\begin{subfigure}{.7\textwidth}
  \centering
  \includegraphics[trim=16mm 0 15mm 10mm,clip,scale=0.55]{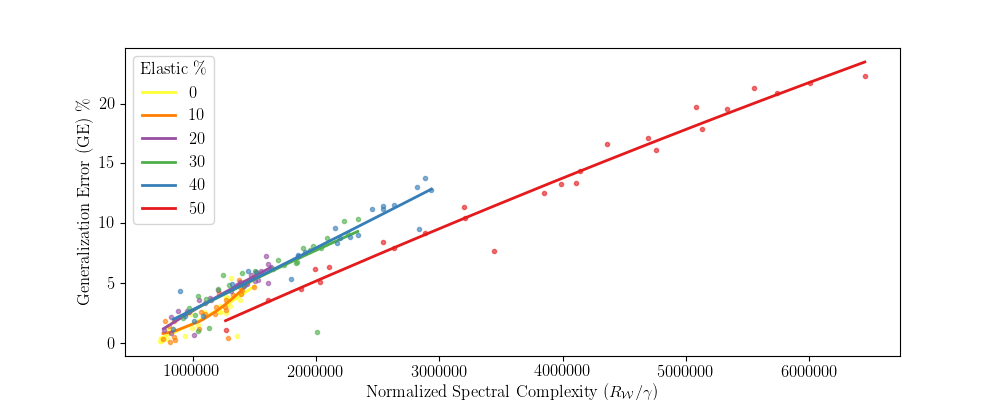}
  \caption{}
  \label{fig:conv_elastic_var_metric}
\end{subfigure}%
\begin{subfigure}{.3\textwidth}
  \centering
  \includegraphics[trim=1mm 0mm 0mm 10mm, clip,scale=0.55]{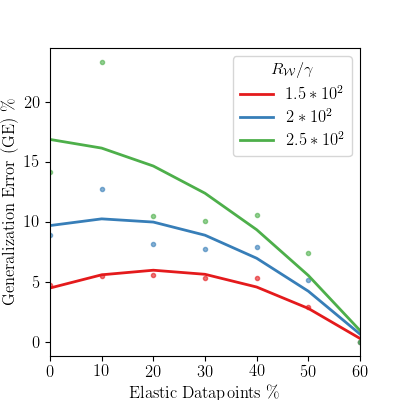}
  \caption{}
  \label{fig:conv_elastic_fixed_metric}
\end{subfigure}
\caption{\textbf{Varying the percentage of translations (a-b) and elastic deformations (c-d)}: We split Training and Testing datasets of constant size into two parts---the first contains images that form a base space, whereas the rest of the dataset contains images that are augmentations of the base space. The percentage values indicate the percentage of the \emph{augmentations} over the total dataset. (a/c) We plot the GE vs spectral complexity. As we increase the number of translations/elastic deformations (equivalently decrease the percentage of the base space) the slopes of the GE curves decrease and we tend to have lower GE for the same spectral complexity metric values. (b/d) We plot the GE vs \% of augmentations for constant complexity values. The percentage of augmentations correlates empirically with the GE indicating that spectral complexity does not account for the architecture invariances.\label{fig:augmentations}}
\end{figure*}

\subsection{Empirical investigation of insensitivity}

We aim to test whether spectral complexity captures accurately the known invariance properties of modern convolutional neural networks. To do this, we increase the relevance of translations and elastic deformations to the image classification task, aiming to give an advantage to invariant architectures. 


Specifically, we created three different versions of the CIFAR-10 dataset: (a) The \textit{control version} consists of 10000 training images and 10000 test images sampled randomly from the CIFAR-10 dataset. (b) The \textit{translated version} is constructed by taking 5000 training images and 5000 test images sampled randomly from the CIFAR-10 dataset. These ``base" sets are then augmented separately with another 5000 images each, that are random translations of the originals. (c) Finally, the \textit{elastic version} is constructed similarly to the translated one, however the base sets are now augmented with images that are random elastic deformations of the originals.

We train using SGD a deep convolutional neural network on each of the above datasets and calculate the GE and the (normalized) spectral complexity metric $R_\mathcal{W}/\gamma$ defined in~\eqref{eq:spectral_complexity} at the end of each epoch. In all following experiments, we used the following architecture: 
\begin{equation}
\begin{split}
&\text{input} \rightarrow 32\text{C}3 \rightarrow \text{MP}2\\ 
&\rightarrow 64\text{C}3 \rightarrow \text{MP}2 \rightarrow 10\text{FC}\rightarrow \text{output},\\
\end{split}
\end{equation}
where $i\text{C}j$ denotes a convolutional layer with $i$ output channels and $j \times j$ filter support, $i\text{FC}$ denotes a fully connected layer with $i$ outputs, and $\text{MP}i$ denotes the max-pooling operator with pooling size of $i$. Our network has $42442$ parameters in total.

Figure~\ref{fig:conv_complete} depicts the GE as a function of the metric for all three datasets, with markers corresponding to results for different epochs. It is important to compare GE values for the \emph{same} spectral complexity as this highlights a hidden variable along which the GE varies that is not captured by spectral complexity alone. We see that for the same metric value the CNN exhibits different GE for the different datasets. The network is able to exploit it's translation invariance and deformation stability to obtain a \emph{lower} GE compared to the normal dataset. Intuitively, by replacing part of the variation in the data manifold with variations to which the network is invariant, we are simplifying the manifold for the CNN improving the GE (even though the complexity of the classifier according to the spectral complexity is the same). We furthermore observe that the CNN is more robust to translations compared to elastic deformations, as it obtains improved GE for former for the same metric values.

To confirm that our results are not specific to the Frobenius norm, but also representative of other spectral complexity definitions, we repeated the experiment also with the (2,1)-norm metric $R_\mathcal{W}'/\gamma$ defined in~\eqref{eq:spectral_complexity_bartlett}. The results were consistent with those presented here and are deferred to the appendix for completeness.

\subsection{Delving deeper into invariances} 

To explore further the insensitivity of spectral complexity to data symmetries, we create datasets with constant size and varying percentage of augmentations. In particular, we start from datasets composed entirely of ``base" samples and gradually increase the percentage of the dataset's augmented images from $0\%$ from to $50\%$. Once more, we create two sets: one with translations and one featuring elastic deformations. We use SGD to train a CNN on these datasets and calculate after each epoch the GE and the spectral complexity metric. 

We plot the results in Figure~\ref{fig:augmentations}. Specifically, Figures~\ref{fig:conv_trans_var_metric} and~\ref{fig:conv_elastic_var_metric} show for the translated and elastic datasets, respectively, that more augmentation results in GE curves that have gradually smaller slopes. Thus, for the same metric, the GE decreases as the number of augmentations increases. Alternatively, we can fix a metric value and plot the GE vs the percentage of normal data-points. We plot the results in Figures~\ref{fig:conv_trans_fixed_metric} and~\ref{fig:conv_elastic_fixed_metric}. We see that, for fixed metric values, the percentage of augmented data-points, i.e., ones that are translations or deformations of others, correlates empirically with the GE. These findings reinforce our hypothesis: spectral complexity is insensitive to the well-known invariances of CNNs and is therefore likely to lead to sub-optimal generalization bounds. 


\section{Comparing convolutional and locally connected networks}

This section aims to provide theoretical evidence supporting that spectral complexity analyses are insensitive to invariances of CNNs. To do so, in Sections~\ref{subsec:cnn} and~\ref{sec:limitations}, we derive respectively generalization bounds for deep neural networks with convolutional and locally-connected layers---the latter maintain the sparsity structure, but do not employ weight sharing. The tightness of our derivation is investigated in Section~\ref{sec:tightness}. Strikingly, we find that both convolutional and locally-connected bounds take, up to log factors, the same form. Our result suggests that spectral complexity analyses exploit the sparsity of convolutional filters but not the invariance properties that arise from the stacking of convolutional layers.   

\subsection{Convolutional networks}
\label{subsec:cnn}

Being derived for fully-connected neural networks, norm-based generalization bounds are not specifically adapted to convolutional architectures. Our first order of business is thus to understand how much one may gain by explicitly considering the structure of convolutions in the generalization error derivation.

To this end, we first aim to tighten the bound of \cite{neyshabur2017pac} and adapt it to the convolutional case. Specifically we will improve upon the architecture dependent constant $\Psi_f$. We show that for the case of convolutional layers the original value of $\Psi_f=d\sqrt{h}$ is unacceptably high.

Our prove the following generalization bound:

\begin{theorem}\label{conv_layer_theorem}
(Generalization Bound). Let $f_{\bw}:\mathbb{R}^{n} \rightarrow \mathbb{R}^k$ be a $d$-layer network, consisting of $|\cC|$ convolutional layers, $|\cF|$ fully-connected layers, and layer-wise ReLU activations. For any $\gamma,\delta > 0$, with probability at least $1-\delta$ over the training set of size $m$ we have 
$$
L_0(f_{\bw}) \leq 
    \hat{L}_{\gamma}(f_{\bw}) 
    + \tilde{\cO}\left( \frac{B \, \Psi_f \, R_{\mathcal{W}}}{\gamma \sqrt{m} } \right),
$$
with $\|x\|_2 \leq B$ being a uniform bound on the input vectors, $R_{\mathcal{W}}$ is as in~\eqref{eq:spectral_complexity}, and
$$\Psi_f = q\sum_{l \in \cC} \sqrt{b_{l}} + \sum_{l \in \cF} \sqrt{s_l}, $$
Above, $q_l$ and $b_l$ denote respectively the filter support and number of output channels of the $l$-th convolutional layer, and $s_l$ counts the number of non-zero entries of the $l$-th fully-connected layer.
%
%
\end{theorem}

\begin{table*}[t]
\begin{center}
\begin{tabular}{ lcccc  }
  \toprule
   & LeNet-5 & AlexNet & VGG-16 \\ 
  \midrule
  \citep{neyshabur2017pac} & $\frac{10^{2.5}}{\sqrt{m} }$ & $\frac{10^{3.5}}{\sqrt{m}}$ & $\frac{10^{4}}{\sqrt{m}}$ \T\\   
  Ours  & $\frac{10^2 }{ \sqrt{m }}$ & $\frac{10^{2.5} }{ \sqrt{m }}$ & $\frac{10^{2.5} }{ \sqrt{m }}$  \\
  \bottomrule
\end{tabular}
\end{center}
\vskip -0.1in
\caption{The value of the generalization error bound $((\Psi_f R_{\mathcal{W}})/(\gamma\sqrt{m}))$ for common feed-forward architectures. For simplicity, we consider a best-case scenario and assume $\Psi_f \approx \gamma \approx 1$.
\vspace{0mm}} 
\label{GE-table}
\end{table*}

The theorem associates the generalization capacity of a deep convolutional neural network to its weights, as well as to key aspects of its architecture. Interestingly, there is a sharp contrast between convolutional and fully-connected layers.

\emph{Fully-connected layers}, in accordance to previous analyses, exhibit a sample complexity that depends linearly on the number of neurons---subject to sparsity constrains that is. For instance, when all layers are sparse with sparsity $s$ and constant stable-rank, ignoring log factors our bound implies that $m = \tilde{\cO}(d^2 s)$ suffice to attain good generalization. For the same setting, the sample complexity was determined as $m = \tilde{\cO}(d^2\max(n_1,n_2))$ by~\cite{neyshabur2017pac}.

\emph{Convolutional layers} contribute more mildly to the sample complexity, with the latter increasing linearly on the filter support $q_l$ and channels $b_l$, but being independent on the layer input size. A case in point, in a fully convolutional network of $d$ layers, each with constant stable-rank and $b_l = b$ output channels, our bound scales like $\cO(q^2 d^2 b)$, while previously it scaled like $\cO(n^2 d^2 b)$. The latter constitutes a two order-of-magnitude improvement when the filter support is $q = O(1)$ (as is usually the case). 


To illustrate these differences resulting from $\Psi_f$, we conduct an experiment on LeNet-5 for the MNIST dataset, and on AlexNet and VGG-16 for the Imagenet dataset. We omit term $R_{\cW} \approx 1$ assuming that $||\bW_l||_F \approx ||\bW_l||_2 \approx 1 , \; \forall i \in \{1,...,K\}$. We plot the results in Table \ref{GE-table}. 
It can be seen that the proposed bounds are orders of magnitude tighter than the previous PAC-Bayesian approach.
%

Clearly, the assumption $R_\cW \approx 1$ is unrealistic in practice. For values obtained by trained networks the bounds presented above are still vacuous by several orders of magnitude. We will see that this looseness has consequences when comparing the bound to the one for locally-connected architectures.

\subsubsection{Proof outline of Theorem \ref{conv_layer_theorem}}

We begin by presenting two prior results which will be useful later. The first relates the noise robustness to perturbations of a classifier to the GE. The second quantifies the perturbation robustness of general deep neural networks. We then outline how these apply to the convolutional setting.
Before proceeding, we recall that, given two probability measures $P$ and $Q$ over a set $X$, the Kullback-Leibler divergence is defined as $\text{KL}(P||Q):=\int_{X}\log \frac{dP}{dQ}dP$.


\paragraph{Useful previous results.}
Let $f_{\bw}$ be any deterministic predictor (not necessarily a neural network). 
The following lemma from \cite{neyshabur2017pac} introduces the condition $\mathbb{P}_{\bu}[\max_{\boldsymbol{x \in \mathcal{X}}} |f_{\bw+\bu }(\bx)-f_{\bw}(\bx)|_2 \leq \frac{\gamma}{4} ]$ as a probabilistic bound on the Lipschitz constant of the predictor $f_{\bw}$, and relates it to the generalization error: 
\begin{lemma}[\cite{neyshabur2017pac}] Let $f_{\bw}(\bx):\mathcal{X} \Rightarrow \mathbb{R}^k$ be any predictor (not necessarily a neural network) with parameters $\bw$, and $P$ be any distribution on the parameters that is independent of the training data. Then, for any random perturbation $\bu$ s.t. $\mathbb{P}_{\bu}[\max_{\boldsymbol{x \in \mathcal{X}}} |f_{\bw+\bu }(\bx)-f_{\bw}(\bx)|_2 \leq \frac{\gamma}{4} ] \geq \frac{1}{2}$, we have 
\begin{equation}
L_0(f_{\bw}) \leq \hat{L}_{\gamma}(f_{\bw})+\cO(\sqrt{\frac{ \normalfont{\text{KL}}(\bw+\bu||P) +\text{ln} \frac{6 m }{\delta} }{ m-1  }}) 
\end{equation}
with probability at least $1-\delta$, 
where $\gamma$ and $\delta$ are positive constants.
\label{lemma:neyshabur-ge}
\end{lemma}

A trade-off can be observed between the condition $\mathbb{P}_{\bu}[\max_{\boldsymbol{x \in \mathcal{X}}} |f_{\bw+\bu }(\bx)-f_{\bw}(\bx)|_2 \leq \frac{\gamma}{4} ] \geq \frac{1}{2}$ and the $\text{KL}$ term in the right hand side of the above inequality. The KL term is \emph{inversely} proportional to the variance of the noise $u$. 
Therefore one would want to maximize the variance of the noise, however the distance $\max_{\boldsymbol{x \in \mathcal{X}}} |f_{\bw+\bu }(\bx)-f_{\bw}(\bx)|_2$ can potentially grow unbounded with high probability for high enough values of the variance. 


Characterizing the condition  $\mathbb{P}_{\bu}[\max_{\boldsymbol{x \in \mathcal{X}}} |f_{\bw+\bu }(\bx)-f_{\bw}(\bx)|_2 \leq \frac{\gamma}{4} ] \geq \frac{1}{2}$ entails understanding the sensitivity of our deep convolutional neural network classifier on random perturbations to the weights. To that end, we review here a useful perturbation bound from \cite{neyshabur2017pac} on the output of a general deep neural network:

\begin{lemma}
[Perturbation bound by~\cite{neyshabur2017pac}] For any $B,d > 0$, let $f_w:\mathcal{X}_{B,n} \Rightarrow \mathbb{R}^k$ be a d-layer network with ReLU activations.. Then for any $\bw$, and $\bx \in \mathcal{X}_{B,n}$, and perturbation $\bu = \text{vec}(\{\bU_l \}^d_{i=1} )$ such that $||\bU_l ||_2 \leq \frac{1}{d}||\bW_l ||_2$, the change in the output of the network can be bounded as follows
\begin{equation}
  |f_{\bw+\bu }(\bx)-f_{\bw}(\bx)|_2 \leq e^2B \tilde{\beta}^{d-1} \sum_l ||\bU_l||_2,
\end{equation}
where $e$, $B$ and $\tilde{\beta}^{d-1}$ are considered as constants after an appropriate normalization of the layer weights. 
\end{lemma}

\begin{figure*}[t!]
\centering
\begin{subfigure}{.5\textwidth}
  \centering
  \includegraphics[scale=0.53]{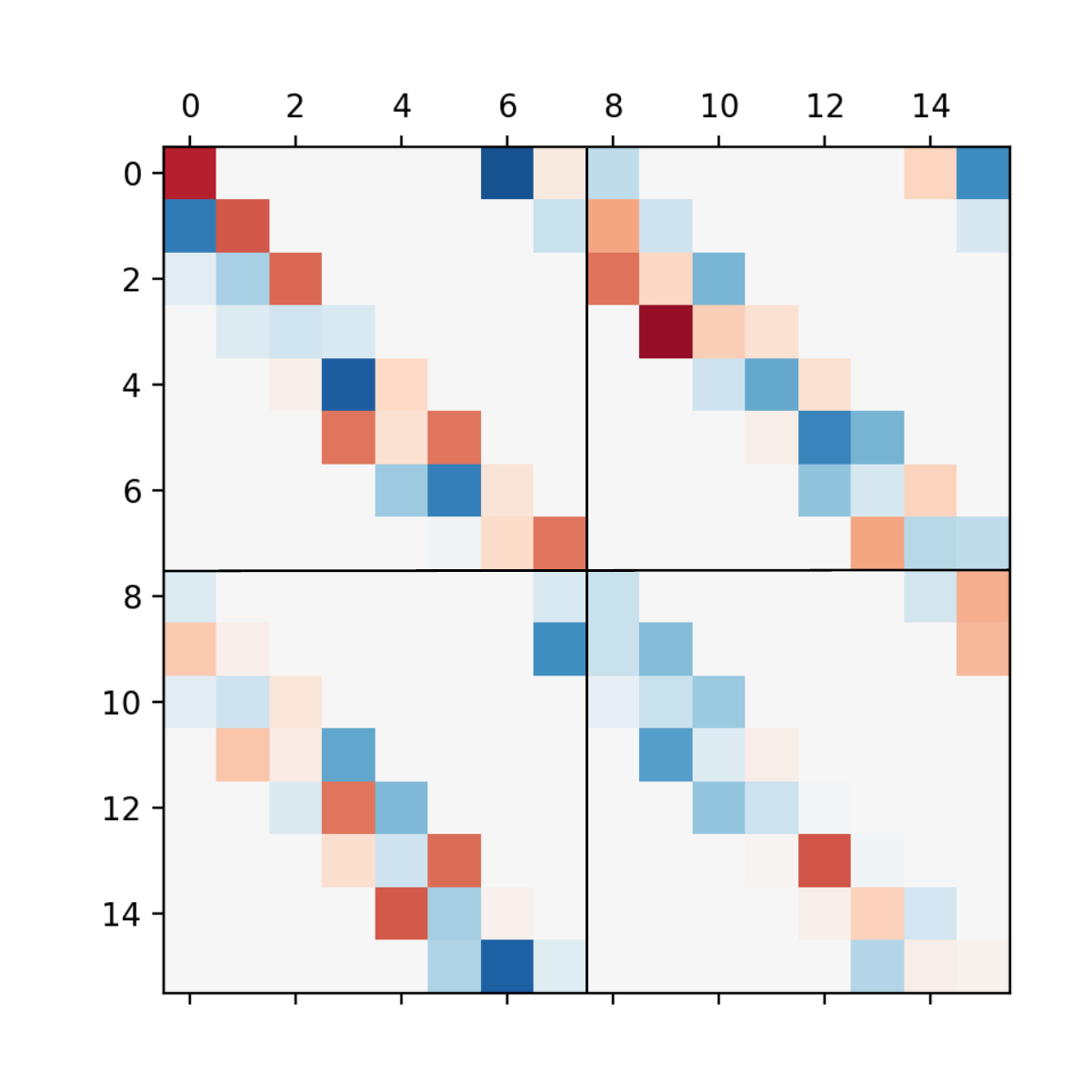}
  \caption{}
  \label{fig:Locally_connected_arch}
\end{subfigure}%
\begin{subfigure}{.5\textwidth}
  \centering
  \includegraphics[scale=0.53]{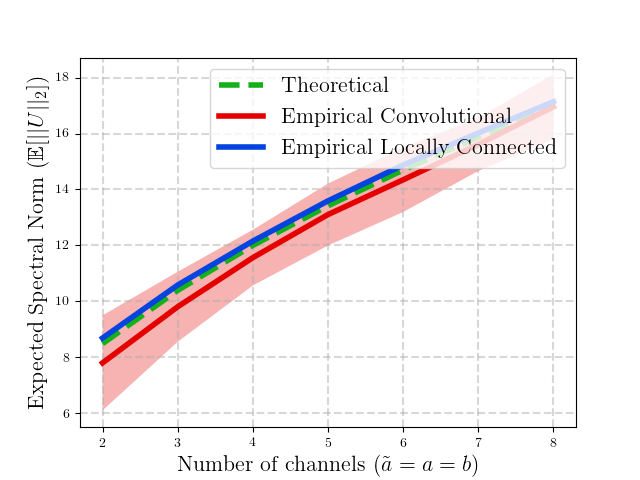}
  \caption{}
  \label{fig:tightness_of_bounds}
\end{subfigure}
\caption{(a) The weight matrix of a 1D locally-connected layer with two input and two output channels. 
(b) We plot empirical and theoretical estimates of the mean $\mathbb{E}[\|\bU\|_2]$. Our theoretical upper estimates (in green) closely follow the empirical estimates for both the convolutional (in red) and the locally-connected (in blue) case. Note that the theoretical estimate is identical for both cases, so we only plot it once. For the empirical estimates, we also show one standard deviation confidence intervals. The locally connected case is much more concentrated than the convolutional and the corresponding confidence interval is not visible in the figure.}
\label{fig:Locally_connected}
\end{figure*}

We note that correctly estimating the spectral norm of the perturbation at each layer is critical to obtaining a tight bound. Specifically, if we exploit the structure of the perturbation we can \emph{increase significantly} the variance of the added perturbation for which  $\mathbb{P}_{\bu}[\max_{\boldsymbol{x \in \mathcal{X}}} |f_{\bw+\bu }(\bx)-f_{\bw}(\bx)|_2 \leq \frac{\gamma}{4} ] \geq \frac{1}{2}$ holds. 

\paragraph{New results.}
The analysis for the convolutional case is difficult due to the fact that the noise per pixel is not independent. We defer the proof to the appendix. We obtain the following lemma, where log parameters have been omitted for clarity:
\begin{lemma}
Let $\bU \in \mathbb{R}^{d_2 \times d_1}$ be the perturbation matrix of a 2$d$ convolutional layer with $a$ input channels, $b$ output channels, convolutional filters $\phi \in \mathbb{R}^{q \times q}$ and feature maps $F \in \mathbb{R}^{N \times N}$. Then, if we vectorize the convolutional filter weights and add a vectorized noise vector $\bu$ such that $\bu \sim \mathcal{N}(0,\sigma^2 \boldsymbol{I}_{abq^2})$, we have
\begin{equation}
||\bU||_2 \leq  \sigma(q [\sqrt{a} + \sqrt{b}] + \sqrt{2\log(\frac{2N^2}{\delta})}), 
\end{equation}
with probability at least $1-\delta$. 
\label{lemma:conv-u}
\end{lemma}
We see that the spectral norm of the noise is independent of the dimensions of the latent feature maps, but it is a function of the root of the filter support $q$, the number of input channels $a$ and the number of output channels $b$.  

With this in place, the following lemma identifies the maximum value of the variance parameter $\sigma^2$ that balances the noise sensitivity with the KL term dependence. 

\begin{lemma}
(Perturbation Bound). For any $B,d > 0$, let $f_w:\mathcal{X}_{B,n} \Rightarrow \mathbb{R}^k$ be a d-layer network with ReLU activations and we denote by $\mathcal{C}$ the set of convolutional layers and $\mathcal{D}$ the set of fully connected layers. Then for any $\bw$, and $\bx \in \mathcal{X}_{B,n}$, and a perturbation for $\bu \sim \mathcal{N}(0,\sigma^2 \boldsymbol{I})$, for any $\gamma > 0$ with 
\begin{equation}
\sigma  = \frac{\gamma}{42 B \tilde{\beta}^{d-1} [\sum_{l \in \mathcal{C}} K_l + \sum_{l \in \mathcal{D}} J_l]}  
\end{equation}
we have
$
  \mathbb{P}_{\bu}[\max_{\boldsymbol{x \in \mathcal{X}}} |f_{\bw+\bu }(\bx)-f_{\bw}(\bx)|_2 \leq \frac{\gamma}{4} ] \geq \frac{1}{2},
$
where $e$, $B$, $\tilde{\beta}^{d-1}$ are considered as constants after an appropriate normalization of the layer weights,
$
K_l = q_l \{\sqrt{a_l}+\sqrt{b_l}+\sqrt{2\log(4N_l^2d)} \}
$
and
$
J_l = q_l \{2\sqrt{s_l}+\sqrt{2\log(2d)} \}.
$
\end{lemma}

Theorem \ref{conv_layer_theorem} follows directly from calculating the KL term in Lemma~\ref{lemma:neyshabur-ge}, by noting that $\bw+\bu \sim \mathcal{N}(\bw,\sigma^2\boldsymbol{I})$, $P \sim \mathcal{N}(0,\sigma^2\boldsymbol{I})$, and that then $\text{KL}(\bw+\bu||P)\leq\frac{|\bw|^2}{2\sigma^2}$.

\subsection{Locally-connected networks}
\label{sec:limitations}

The improvement we attained by taking into account the structure of convolutional layers, though significant, still falls short from explaining why deep CNNs are able to generalize beyond the training set---the bounds are too pessimistic. 

Locally-connected layers have a sparse banded structure similar to convolutions, with the simplifying assumption that the weights of the translated filters are not shared. The weight matrix is exemplified in Figure~\ref{fig:Locally_connected_arch} for the case of one-dimensional convolutions. While this type of layer is not used in practice, it enables us to isolate the effect of sparsity on the generalization error.
We prove the following:

\begin{theorem}\label{loc_layer_theorem}
Let $f_{\bw}:\mathbb{R}^{n} \rightarrow \mathbb{R}^k$ be a $d$-layer network, consisting of $|\cL|$ locally-connected layers, $|\cF|$ fully-connected layers, and layer-wise ReLU activations. For any $\gamma,\delta > 0$, with probability at least $1-\delta$ over the training set of size $m$ we have 
$$
L_0(f_{\bw}) \leq 
    \hat{L}_{\gamma}(f_{\bw}) 
    + \tilde{\cO}\left( \frac{B \, \Psi_f \, R_{\mathcal{W}}}{\gamma \sqrt{m} } \right),
$$
with $\|x\|_2 \leq B$ being a uniform bound on the input vectors, $R_{\mathcal{W}}$ is as in~\eqref{eq:spectral_complexity}, and
$$\Psi_f = q\sum_{l \in \cC} \sqrt{b_{l}} + \sum_{l \in \cF} \sqrt{s_l}. $$
Above, $q_l$ and $b_l$ denote respectively the filter support and number of output channels of the $l$-th locally-connected layer, and $s_l$ counts the number of non-zero entries of the $l$-th fully-connected layer.
%
%
\end{theorem}

Surprisingly for the given choice of spectral complexity the obtained bounds for convolutional and locally connected layers are identical up to log factors that are artifacts of the derivation. Implicitly, the hypothesis class $\mathcal{H}$ induced by spectral complexity is large enough to include both convolutional and non-convolutional architectures. At the same time, the bounds in both cases hold for \emph{any} data distribution $\mathcal{D}$. These two points stand in stark contrast with common design practice where the hypothesis class $\mathcal{H}$ is assumed to be convolutional architectures, with good generalization properties specifically for data distributions $\mathcal{D}$ that represent natural images.

In hindsight, it might be clear that the upper bound on convolutional layers is not tight. However, all of the above are not evident in previous analyses and consequently misleading conclusions can be drawn when validating bounds through simple empirical correlation.

\subsubsection{Proof outline of Theorem~\ref{loc_layer_theorem}}

The analysis is similar to the case of convolutional layers, with the exception of 
how term $\|\bU\|_2$ is bounded (see Lemma~\ref{lemma:conv-u}). We now have:
\begin{lemma}
Let $\bU \in \mathbb{R}^{d_2 \times d_1}$ be the perturbation matrix of a 2$d$ locally-connected layer with $a$ input channels, $b$ output channels, filters $\phi \in \mathbb{R}^{q \times q}$ and feature maps $F \in \mathbb{R}^{N \times N}$. Then if non-zero elements follow $\bU_{i,j} \sim \mathcal{N}(0,\sigma^2)$, we have
\begin{equation}
||\bU||_2 \leq  \cO(\sigma(q [\sqrt{a} + \sqrt{b}] + \sqrt{2\log(\frac{1}{\delta})})),
\end{equation}
with probability at least $1-\delta$.
\end{lemma}
%
The rest of the proof technique is identical to that used for convolutional layers and is deferred to the appendix.

\subsection{Empirical investigation of tightness}
\label{sec:tightness}
Theorems \ref{conv_layer_theorem} and \ref{loc_layer_theorem} depend on $\mathbb{E}[||\bU||_2]$, the expected spectral norm of the layer noise.  We test our concentration bounds by computing analytically and empirically $\mathbb{E}[||\bU||_2]$ for synthetic data. 

Our experiment considers 1D signals, filters $\phi \in \mathbb{R}^{9}$, feature maps $F \in \mathbb{R}^{100}$, $a$ input channels, $b$ output channels. We calculate the spectral norm $||\bU||_2$ while increasing the number of input and output channels with $\tilde{a} := a = b$. To obtain empirical estimates, we average the results over $N=100$ iterations for each choice of $\tilde{a}$. As seen in Figure~\ref{fig:tightness_of_bounds}, the theoretical values closely match the empirical estimates. 

\section{Discussion}

Two recent works \citep{kawaguchi2017generalization, nagarajan2019uniform} have discussed limitations of uniform convergence in explaining generalization in deep learning. They derive counter examples in constrained settings were uniform convergence provably cannot explain generalization. Our results can be cast in the same light. Implicitly we have shown that the hypothesis class induced a posteriory by spectral complexity includes elements with greatly varying generalization error, and in specific cases includes both convolutional and non-convolutional architectures. Crucially we have focused on the fact that there are no assumptions on the data distribution in current analyses. 
We believe that incorporating invariances of deep neural networks in future analyses is a crucial component of non-vacuous generalization error bounds. However, this does not directly address the issue of uniform convergence which might require moving to the analysis of single hypotheses.

\clearpage
\newpage


\subsubsection*{References}

\bibliographystyle{plainnat}
\bibliography{aistats2020}

\onecolumn

\title{Appendix for: ``The role of invariance in spectral complexity-based generalization bounds''
}
\date{}
\maketitle

We denote vectors with bold lowercase letters and matrices with bold capital letters. Given two probability measures $P$ and $Q$ over a set $X$ we define the Kullback-Leibler divergence as $\text{KL}(P||Q)=\int_{X}\log \frac{dP}{dQ}dP$. We denote with $\varphi(x) = \frac{1}{\sqrt{2\pi}}e^{-\frac{1}{2}x^2}$ the Gaussian kernel.

\section{Detailed proof of Theorem 4.1}

In the derivations below we will rely upon the following useful theorem for the concentration of the spectral norm of sparse random matrices 

\begin{customthm}{1.1}\label{bandeira}
\cite{bandeira2016sharp} Let $\boldsymbol{A}$ be a $d_2 \times d_1$ random rectangular matrix with $\boldsymbol{A}_{ij} = \xi_{ij} \psi_{ij}$ where $ \{ \xi_{ij}:1 \leq i \leq d_2 , 1\leq j \leq d_1 \} $ are independent $ \mathcal{N}(0,1)$ random variables and $\{ \psi_{ij}:1 \leq i \leq d_2 , 1\leq j \leq d_1 \}$ are scalars. Then

\begin{equation}
\mathbb{P}(||\boldsymbol{A}||_2 \geq (1+\epsilon) \{ \sigma_1 + \sigma_2 + \frac{5}{\sqrt{\log (1+\epsilon)} }\sigma_* \sqrt{\log (\max(d_2,d_1) )}  + t  \} ) \leq e^{-t^2 / 2 \sigma_*^2 }
\end{equation}

for any $0 \leq \epsilon \leq 1/2$ and  $t \geq 0$ with

\begin{equation}
\sigma_1:= \max_i \sqrt{ \sum_j \psi_{ij}^2 } \qquad \sigma_2:= \max_i \sqrt{ \sum_j \psi_{ij}^2 } \qquad \sigma_*:= \max_{ij}|\psi_{ij}|.
\end{equation}
\end{customthm}

In the following we will use the same numbering for Theorems and Lemmas as in the main paper. Theorems and Lemmas unique to the appendix will be numbered with a prefix corresponding to the section where the theorem is introduced and suffix with a corresponding number.

\subsection*{A. Fully Connected Layers}

\begin{customlem}{A.1}\label{fully_connected_spectral}
Let $\boldsymbol{U} \in \mathbb{R}^{d_2 \times d_1}$ be the perturbation matrix of a fully connected layer with with row and column sparsity equal to $s$. Then if non-zero elements follow $\boldsymbol{U}_{i,j} \sim \mathcal{N}(0,\sigma^2)$, with probability greater than $1-\delta$
\begin{equation}
||\boldsymbol{U}||_2 \leq \mathcal{O}(\sigma ( 2\sqrt{s} + \sqrt{2\log(\frac{1}{\delta})})).
\end{equation}
%
\end{customlem}

\begin{proof}
For $\boldsymbol{u} \sim \mathcal{N}(0, I)$ we need define an index function that allows a Gaussian random noise variable at the locations where the original dense layer is non-zero. 

We assume that $\psi_{ij}=1$ when $\boldsymbol{U}_{ij} \ne 0$ and is zero otherwise, and get the result trivially from Theorem \ref{bandeira}. We can extend the result to $\sigma > 0$ by considering that $||\sigma \boldsymbol{U}_l ||_2 = \sigma||\boldsymbol{U}_l ||_2$.
\end{proof}

\begin{figure*}[t!]
\centering
\begin{subfigure}{.5\textwidth}
  \centering
  \includegraphics[scale = 0.5]{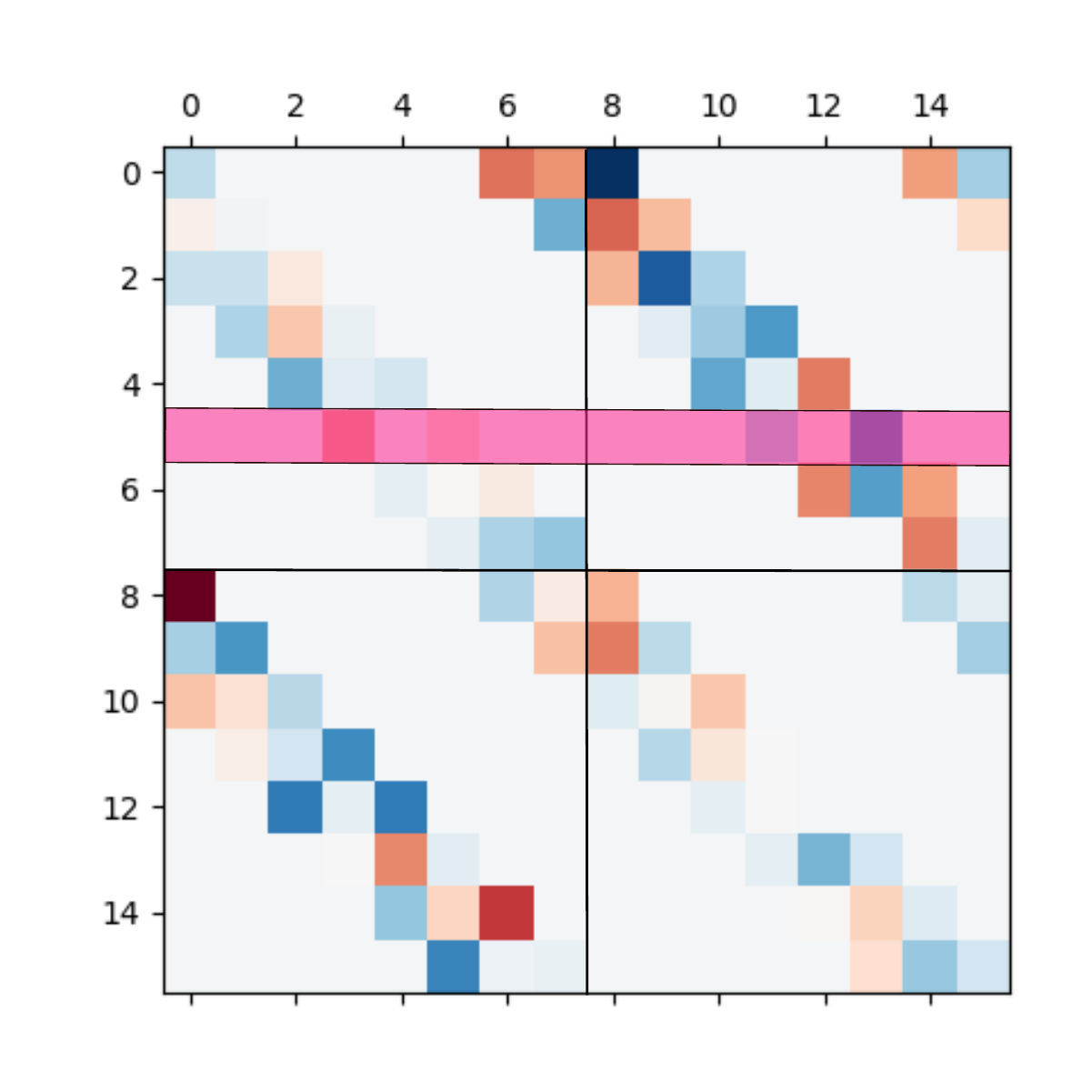}
  \caption{$\sigma_1:= \max_i \sqrt{ \sum_j \psi_{ij}^2 }$ }
  \label{fig:one_dim_loc_row}
\end{subfigure}%
\begin{subfigure}{.5\textwidth}
  \centering
  \includegraphics[scale = 0.5]{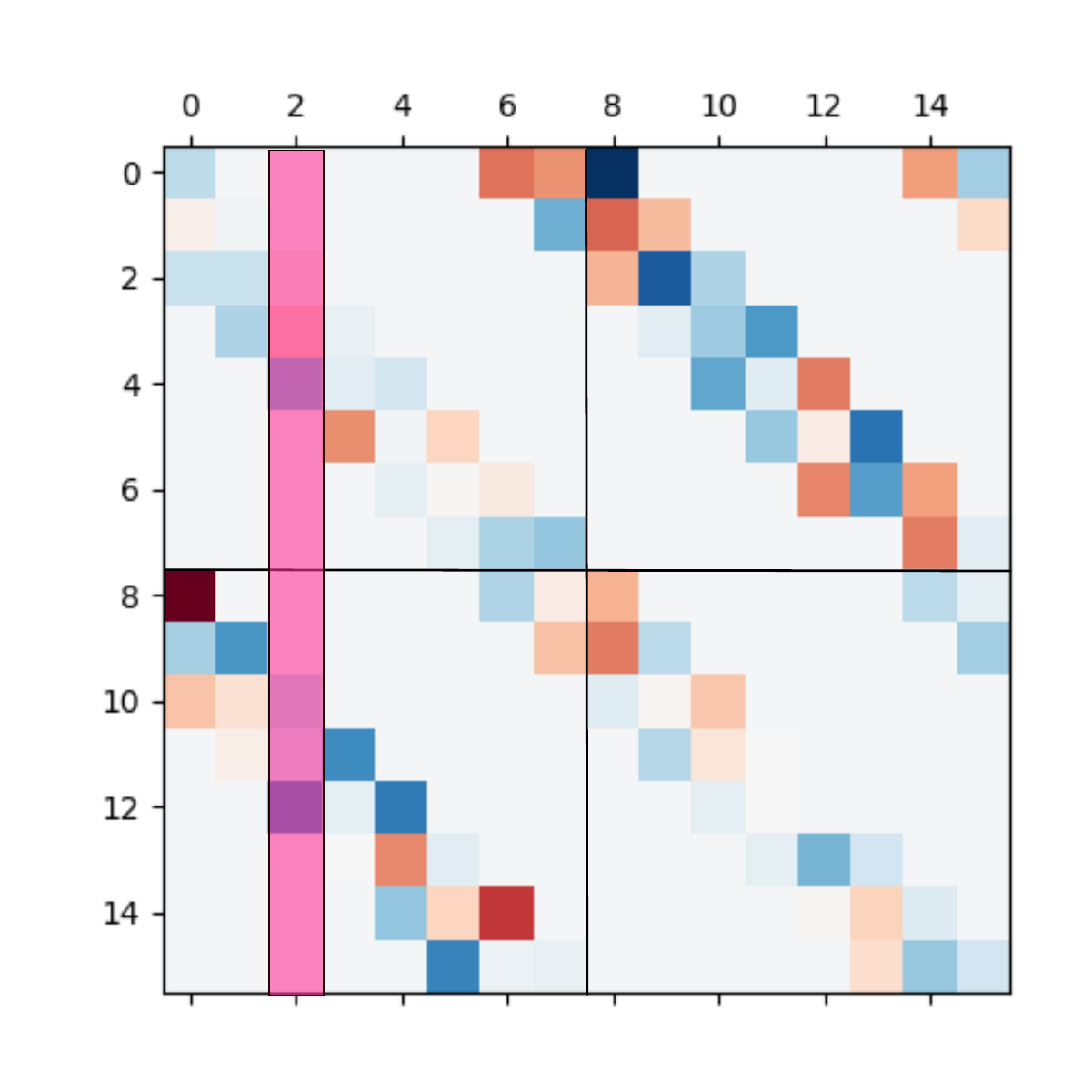}
  \caption{$\sigma_2:= \max_j \sqrt{ \sum_i \psi_{ij}^2 }$}
  \label{fig:one_dim_loc_col}
\end{subfigure}
\caption{$\sigma_1$ and $\sigma_2$}
\label{fig:one_dim_loc}
\end{figure*}

\subsection*{B. Locally Connected Layers}

\begin{customlem}{4.7}
Let $\boldsymbol{U} \in \mathbb{R}^{d_2 \times d_1}$ be the perturbation matrix of a 2$d$ locally connected layer with $a$ input channels, $b$ output channels, filters $\phi \in \mathbb{R}^{q \times q}$ and feature maps $F \in \mathbb{R}^{N \times N}$. Then, if non-zero elements follow $\boldsymbol{U}_{i,j} \sim \mathcal{N}(0,\sigma^2)$, we have
\begin{equation}
||\boldsymbol{U}||_2 \leq  \mathcal{O}(\sigma(q [\sqrt{a} + \sqrt{b}] + \sqrt{2\log(\frac{1}{\delta})})), 
\end{equation}
with probability greater than $1-\delta$.
\end{customlem}

\begin{figure*}[t!]
\includegraphics[scale = 0.3]{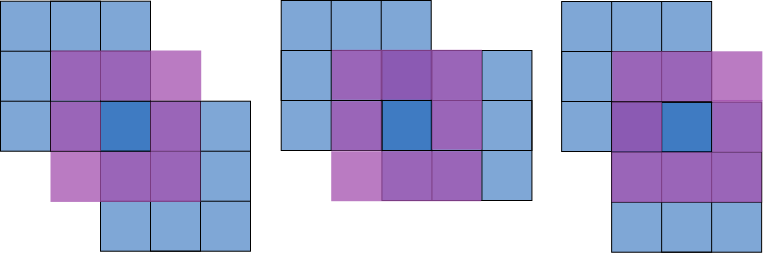}
\centering
\caption{Possible shifts with overlap: With blue we plot a 2d filter $f \in \mathbb{R}^{3\times3}$ and 3 filters $f \in \mathbb{R}^{3\times3}$ that overlap with it's bottom right pixel. In purple we plot the box denoting the boundaries of the set of all shifted filters that overlap with the bottom right pixel.} 
\label{fig:shifts}
\end{figure*}

\begin{proof}
We will consider first the case $\boldsymbol{u} \sim \mathcal{N}(0, I)$. A convolutional layer is characterised by it's output channels. For each output channel each input channel is convolved with an independent filter resulting in a set of feature maps. For each output channel these feature maps are then summed together. We consider locally connected layers, i.e. the layers are banded in the same way as convolutions but the entries are independent and there is no weight sharing. For the case of one dimensional signals the implied structure is plotted in Figure \ref{fig:one_dim_loc}. 

Similar to Lemma A.1 we assume that $\psi_{ij}=1$ when $\boldsymbol{U}_{ij} \ne 0$ and is zero otherwise. We need to evaluate $\sigma_1:= \max_i \sqrt{ \sum_j \psi_{ij}^2 }$ , $\sigma_2:= \max_j \sqrt{ \sum_i \psi_{ij}^2 }$ and $\sigma_*:= \max_{ij}|\psi_{ij}|$ for a matrix like the one in Figure \ref{fig:one_dim_loc}. 

We plot what these sums represent in Figures \ref{fig:one_dim_loc_row}, \ref{fig:one_dim_loc_col}. We are however working typically with 2 dimensional signals. For $\sigma_1$ we can find an upper bound, by considering that the sum for a given filter and a given pixel location represents the maximum number of overlaps for all 2d shifts. For the case of 2d this is $q^2$, equal to the support of the filters. We plot these shifts in Figure \ref{fig:shifts}. We also need to consider that there are $a$ input channels. We then get 
\begin{equation}
\sigma_1 := \max_i \sqrt{ \sum_j \psi_{ij}^2 } =  \sqrt{\sum_{a}\sum_{q^2} 1^2 }  =  \sqrt{aq^2} =  q\sqrt{a}  
\end{equation}
For $\sigma_2$ each column in the matrix represents a concatenation of convolutional filters $f \in \mathbb{R}^{q \times q} $. The support of the filters is $q^2$ and there are $b$ filters stacked on top of eachother, corresponding to the $b$ output channels. Then it is straight forward to derive that
\begin{equation}
\sigma_1 := \max_i \sqrt{ \sum_j \psi_{ij}^2 } = \sqrt{\sum_{b}\sum_{q^2} 1^2 }  =  \sqrt{bq^2}  =  q\sqrt{b}.  
\end{equation}
Furthermore trivially $\sigma_* = 1$ and when $\sigma > 0$ we can get the result by considering that $||\sigma \boldsymbol{U}_l ||_2 = \sigma||\boldsymbol{U}_l ||_2$.
\end{proof}

\subsection*{C. Convolutional Layers}

\begin{customlem}{4.4}\label{spectral_norm_for_convolutions}
Let $\boldsymbol{U} \in \mathbb{R}^{d_2 \times d_1}$ be the perturbation matrix of a 2$d$ convolutional layer with $a$ input channels, $b$ output channels, convolutional filters $\phi \in \mathbb{R}^{q \times q}$ and feature maps $F \in \mathbb{R}^{N \times N}$. Then, if we vectorize the convolutional filter weights and add a vectorized noise vector $\boldsymbol{u}$ such that $\boldsymbol{u} \sim \mathcal{N}(0,\sigma^2 \boldsymbol{I}_{abq^2})$, we have
\begin{equation}
||\boldsymbol{U}||_2 \leq  \sigma(q [\sqrt{a} + \sqrt{b}] + \sqrt{2\log(\frac{2N^2}{\delta})}),
\end{equation}
with probability greater than $1-\delta$.
\end{customlem}

\begin{proof}

We consider noise filters $f \in \mathbb{R}^{q \times q}$ and feature maps $F \in \mathbb{R}^{N \times N}$. We define the convolutional noise operator from input channel $j$ to output channel $i$ in the spatial domain as $\boldsymbol{A}^{ij} \in \mathbb{R}^{N^2 \times N^2} $ and in the frequency domain as $\boldsymbol{\tilde{A}}^{ij} \in \mathbb{C}^{N^2 \times N^2}$ and we denote the Fourier transform matrix as $\boldsymbol{F} \in \mathbb{C}^{N^2 \times N^2}$. Each convolutional operator corresponds to one convolutional noise filter $f^{ij} \in \mathbb{R}^{q \times q}$. We can now define the structure of the 2d convolutional noise matrix $\boldsymbol{U}$. Given $a$ input channels and $b$ output channels the noise matrix $\boldsymbol{U}$ is structured as
\begin{equation}
\boldsymbol{U} = 
\begin{bmatrix}
    \boldsymbol{A}^{00}  & ... & \boldsymbol{A}^{0a}  \\
    \vdots  & \ddots & \vdots  \\
    \boldsymbol{A}^{b0}  & ... & \boldsymbol{A}^{ba}  \\
\end{bmatrix}
\end{equation}

were for all $b$ output channels the signal's $a$ input channels are convolved with independent noise filters and the results of these convolutions are summed up.

By exploiting the unitary-invariance property of the spectral norm, we transform this matrix into the Fourier domain to obtain
\begin{equation}
\begin{split}
||\boldsymbol{U}||_2 
& =
||
(\boldsymbol{I}_b \otimes \boldsymbol{F}^T)
\begin{bmatrix}
    \boldsymbol{\tilde{A} }^{00}  & ... & \boldsymbol{\tilde{A}}^{0a}  \\
    \vdots  & \ddots & \vdots  \\
    \boldsymbol{\tilde{A}}^{b0}  & ... & \boldsymbol{\tilde{A}}^{ba}  \\
\end{bmatrix}
(\boldsymbol{I}_a \otimes \boldsymbol{F})
||_2  =
||
\begin{bmatrix}
    \boldsymbol{\tilde{A} }^{00}  & ... & \boldsymbol{\tilde{A}}^{0a}  \\
    \vdots  & \ddots & \vdots  \\
    \boldsymbol{\tilde{A}}^{b0}  & ... & \boldsymbol{\tilde{A}}^{ba}  \\
\end{bmatrix}
||_2 = 
||
\begin{bmatrix}
    \boldsymbol{\tilde{B}}_{0}  & ... & 0  \\
    \vdots  & \ddots & \vdots  \\
    0  & ... & \boldsymbol{\tilde{B}}_{N^2}  \\
\end{bmatrix} ||_2, 
\end{split}
\end{equation}
where we have used the fact that the matrices $\boldsymbol{\tilde{A}}^{ij}$ are diagonal and a concatenation of diagonal matrices can always be rearranged into block diagonal form. In our case, we have defined blocks
\begin{equation}
\boldsymbol{\tilde{B}_{n}} = 
\begin{bmatrix}
    \lambda_n^{00}  & \hdots & \lambda_n^{0a}  \\
    \vdots  & \ddots & \hdots  \\
    \lambda_n^{b0}  & \hdots & \lambda_n^{ba}  \\
\end{bmatrix}
\end{equation}
with entries
\begin{figure*}[t]
\centering
\begin{subfigure}{.5\textwidth}
  \centering
  \includegraphics[scale=0.5]{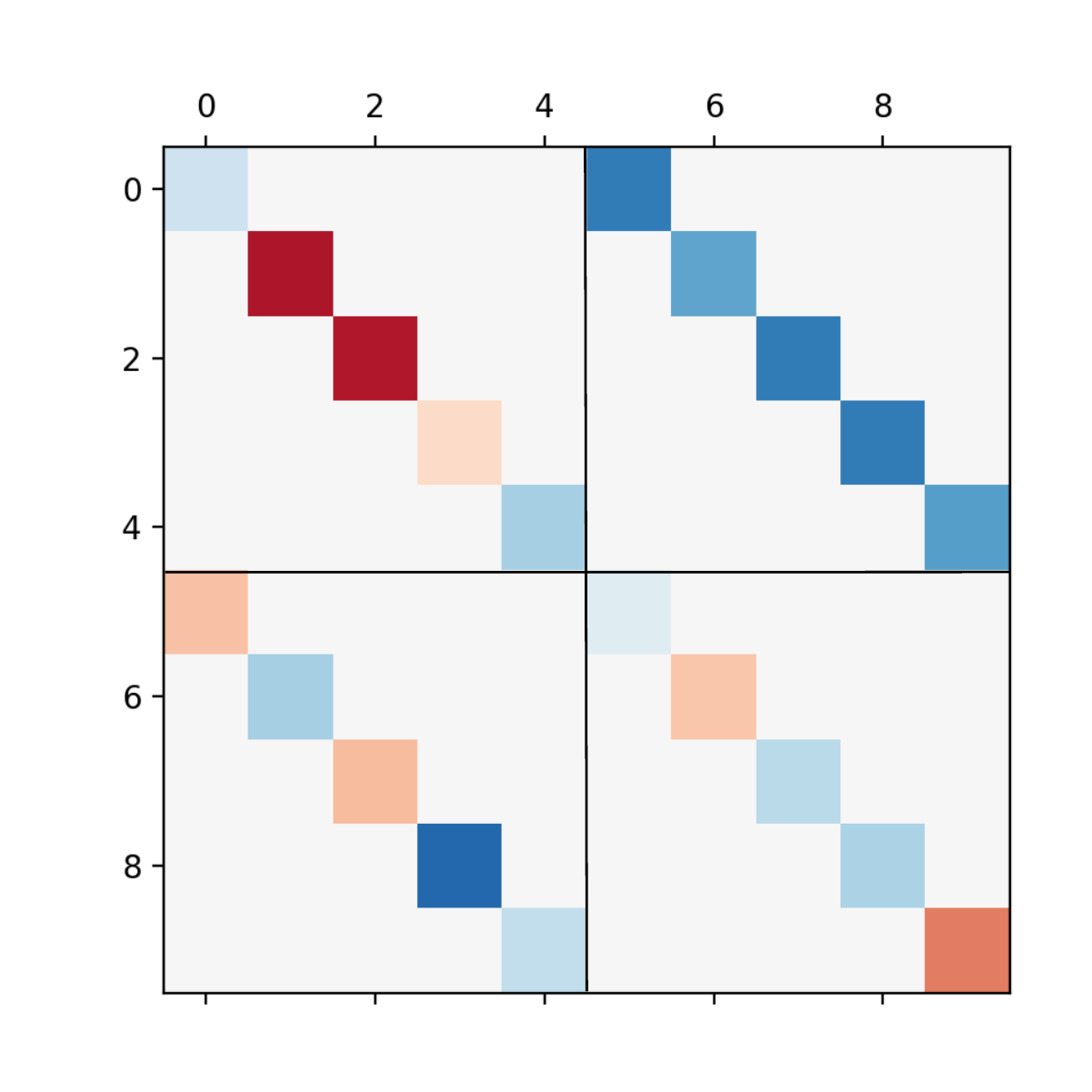}
  \caption{}
\end{subfigure}%
\begin{subfigure}{.5\textwidth}
  \centering
  \includegraphics[scale=0.5]{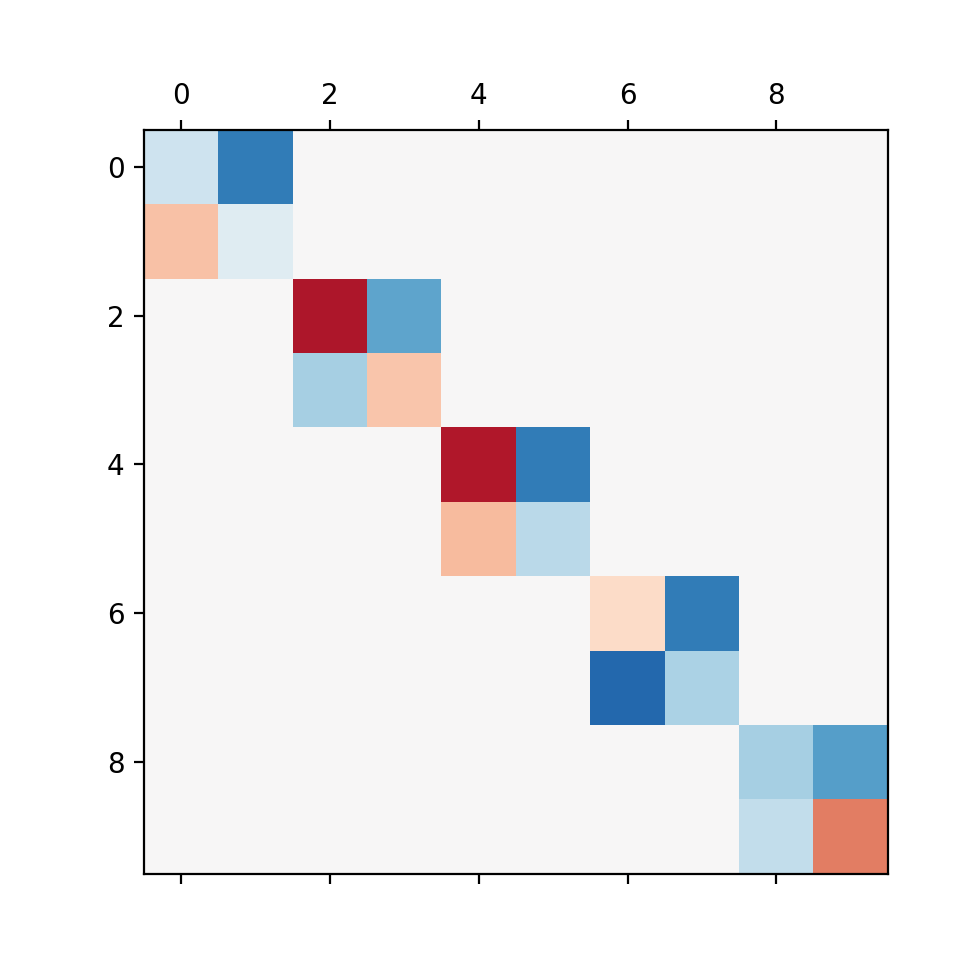}
  \caption{}
\end{subfigure}
\caption{\textbf{Concatenation of diagonal matrices}: We see that the concatenation of diagonal matrices can be always rearranged into a block diagonal matrix.}
\end{figure*}
\begin{equation}
\begin{split}
&\lambda_n^{ij} = \lambda_{n_1,n_2}^{ij} = \sum_{k_1 = 0}^{q - 1} \sum_{k_2 = 0}^{q - 1} e^{-2 \pi i (\frac{k_1 n_1}{q}+\frac{k_2 n_2}{q})} f^{ij}_{k_1,k_2}\\ 
& = \sum_{k_1 = 0}^{q - 1} \sum_{k_2 = 0}^{q - 1} \text{cos}(2 \pi  (\frac{k_1 n_1}{q}+\frac{k_2 n_2}{q})) f^{ij}_{k_1,k_2} + i \sum_{k_1 = 0}^{q - 1} \sum_{k_2 = 0}^{q - 1} \text{sin}(2 \pi (\frac{k_1 n_1}{q}+\frac{k_2 n_2}{q})) f^{ij}_{k_1,k_2},
\end{split}
\end{equation}
where $n_1$,$n_2$ are the frequency coordinates.
In this way the $n^{th}$ block $\boldsymbol{\tilde{B}_{n}}$ corresponds to the $n^{th}$ frequency components from the fourier transforms of all $f^{ij} \;\; \forall i \in \{1,...,b\},\forall i \in \{1,...,a\}$ filters. 

We will also need the matrices $\text{Re}(\boldsymbol{\tilde{B}}_n)$ and $\text{Im}(\boldsymbol{\tilde{B}}_n)$
\begin{equation}
\text{Re}(\boldsymbol{\tilde{B}}_n) = 
\begin{bmatrix}
    \text{Re}(\lambda_n^{00})  & \hdots & \text{Re}(\lambda_n^{0a})  \\
    \vdots  & \ddots & \hdots  \\
    \text{Re}(\lambda_n^{b0})  & \hdots & \text{Re}(\lambda_n^{ba})  \\
\end{bmatrix},
\; \; \;
\text{Im}(\boldsymbol{\tilde{B}}_n) = 
\begin{bmatrix}
    \text{Im}(\lambda_n^{00})  & \hdots & \text{Im}(\lambda_n^{0a})  \\
    \vdots  & \ddots & \hdots  \\
    \text{Im}(\lambda_n^{b0})  & \hdots & \text{Im}(\lambda_n^{ba})  \\
\end{bmatrix}
\end{equation}
The entries of these matrices have the following distributions:
\begin{equation}
\begin{split}
& \text{Re}(\lambda_{n}^{ij}) \sim \mathcal{N}(0, \sigma_{re,n}^2 ) \sim \mathcal{N}(0, \sum_{k_1 = 0}^{q - 1} \sum_{k_2 = 0}^{q - 1}\text{cos}^2(2 \pi  (\frac{k_1 n_1}{q}+\frac{k_2 n_2}{q}))  \\
& \text{Im}(\lambda_{n}^{ij}) \sim \mathcal{N}(0, \sigma_{im,n}^2 ) \sim \mathcal{N} (0, \sum_{k_1 = 0}^{q - 1} \sum_{k_2 = 0}^{q - 1}\text{sin}^2(2 \pi (\frac{k_1 n_1}{q}+\frac{k_2 n_2}{q})),
\end{split}
\end{equation}
where we have used the fact that $f^{ij}_{k_1,k_2}$ are i.i.d Gaussian.

We have now turned our initial problem into a form that lends itself more easily to a solution. Our original matrix has been turned into block diagonal form and each block can be split into real and imaginary parts that have independent gaussian entries, we note however that blocks are not independent of eachother. We will now derive a concentration bound on the original matrix by using the fact that the spectral norm of a block diagonal matrix is equal to the maximum of the spectral norms of the individual blocks.

We can write the following inequalities
\begin{equation}
\mathbb{P}(||\boldsymbol{U}||_2 \leq \epsilon)=\mathbb{P}(\bigcap_{n}\{ ||\boldsymbol{\tilde{B}}_n||_2 \leq \epsilon\}) \geq \mathbb{P}(\bigcap_{n}\{ ||\text{Re}(\boldsymbol{\tilde{B}}_n)||_2 + ||\text{Im}(\boldsymbol{\tilde{B}}_n)||_2 \leq \epsilon\})
\end{equation}
By setting $\epsilon = \max_n(\epsilon_n)$ and $\epsilon_n$ arbitrary constants, we can furthermore write
\begin{equation}
\begin{split}
& \mathbb{P}(||\boldsymbol{U}||_2 \leq \max_n(\epsilon_n)) \\
& \geq \mathbb{P}(\bigcap_{n}\{ ||\text{Re}(\boldsymbol{\tilde{B}}_n)||_2 + ||\text{Im}(\boldsymbol{\tilde{B}}_n)||_2 \leq \max_n(\epsilon_n)\}) \\ 
& \geq \mathbb{P}(\bigcap_{n}\{ ||\text{Re}(\boldsymbol{\tilde{B}}_n)||_2 + ||\text{Im}(\boldsymbol{\tilde{B}}_n)||_2 \leq \epsilon_n\}) \\
& \geq \mathbb{P}(\bigcap_{n} [ \{ ||\text{Re}(\boldsymbol{\tilde{B}}_n)||_2 \leq \epsilon_{n,re} \} \cap \{ ||\text{Im}(\boldsymbol{\tilde{B}}_n)||_2 \leq \epsilon_{n,im}\}] ) \\
& \geq 1 - \sum_{n=1}^{N^2}[\delta_{n,re}+\delta_{n,re}],
\end{split}
\end{equation}
where in line 4 we set $\epsilon_n = \epsilon_{n,re} + \epsilon_{n,im}$ and in line 5 we used a union bound and assumed that $\mathbb{P}(|| \text{Re}(\boldsymbol{\tilde{B}_{n}}) ||_2 \geq \epsilon_{n,re}) \leq \delta_{n,re}$ and $\mathbb{P}(|| \text{Im}(\boldsymbol{\tilde{B}_{n}}) ||_2 \geq \epsilon_{n,im}) \leq \delta_{n,im}$ for positive constants $ \{ \epsilon_{n,re},\epsilon_{n,im},\delta_{n,re},\delta_{n,im} \} \in \mathbb{R}_{+}$.

We will now calculate concentration inequalities for the individual blocks $\text{Re}(\boldsymbol{\tilde{B}}_n)$ and $\text{Im}(\boldsymbol{\tilde{B}}_n)$, turning the general formula we have derived into a specific one for our case. To do that we first apply the following concentration inequality by \cite{vershynin2010introduction}
\begin{customthm}{C.1}
Let $\boldsymbol{A}$ be an $N \times n$ matrix whose entries are independent Gaussian random variables with variance $\sigma^2$. Then for every $t \geq 0$
\begin{equation}
\mathbb{P}(||\boldsymbol{A}||_2 \geq \sigma(\sqrt{N}+\sqrt{n}+\sqrt{2\ln(\frac{1}{\delta} )} )) \leq \delta
\end{equation}
\end{customthm}

on the matrices $\text{Re}(\boldsymbol{\tilde{B}}_n)$ and $\text{Im}(\boldsymbol{\tilde{B}}_n)$. We obtain the following concentration inequalities:
\begin{equation}
\begin{split}
& \mathbb{P}(|| \text{Re}(\boldsymbol{\tilde{B}_{n}}) ||_2 \geq \sigma_{re,n}(\sqrt{a}+\sqrt{b}+\sqrt{2\ln(\frac{1}{\delta_{n,re}} )} )) \leq \delta_{n,re} \\
& \mathbb{P}(|| \text{Im}(\boldsymbol{\tilde{B}_{n}}) ||_2 \geq \sigma_{im,n}(\sqrt{a}+\sqrt{b}+\sqrt{2\ln(\frac{1}{\delta_{n,im}} )} )) \leq \delta_{n,im} \\
\end{split}
\end{equation}
We then make the following calculations which will prove useful:
\begin{equation}
\begin{split}
\max_n [\sigma_{re,n}+\sigma_{im,n}] &= \max_n [\sqrt{ \sum_{k_1 = 0}^{q - 1} \sum_{k_2 = 0}^{q - 1}\text{cos}^2(2 \pi (\frac{k_1 n_1}{q}+\frac{k_2 n_2}{q}) } \\
&+ \sqrt{ \sum_{k_1 = 0}^{q - 1} \sum_{k_2 = 0}^{q - 1}\text{sin}^2(2 \pi (\frac{k_1 n_1}{q}+\frac{k_2 n_2}{q}) } ]\\
& \leq \sqrt{ \sum_{k_1 = 0}^{q - 1} \sum_{k_2 = 0}^{q - 1}\frac{1}{2}} + \sqrt{ \sum_{k_1 = 0}^{q - 1} \sum_{k_2 = 0}^{q - 1}\frac{1}{2} } = \frac{2}{\sqrt{2}}q \leq 1.5q
\end{split}
\end{equation}
since
\begin{equation}
\begin{split}
& \frac{\partial}{\partial\theta_l}(\sqrt{\sum_i\sin^2(\theta_l)}+\sqrt{\sum_i\cos^2(\theta_l)}) = \frac{1}{2}\frac{2\cos(\theta_l)\sin(\theta_l)}{|\sin(\theta_l)|} - \frac{1}{2}\frac{2\cos(\theta_l)\sin(\theta_l)}{|\cos(\theta_l)|} \\
& \frac{\sin(\theta_l)\cos(\theta_l)}{|\sin(\theta_l)||\cos(\theta_l)|}(|\cos(\theta_l)|-|\sin(\theta_l)|) = 0,
\end{split}
\end{equation}
which implies
\begin{equation}
\cos(\theta_l) = \sin(\theta_l) = \pm \frac{1}{\sqrt{2}}.
\end{equation}
We can now substitute $\delta_{n,re} = \delta_{n,im} = \delta/(2N^2), \; \forall n \in \{1,...,N^2\}$ in equation 17. We get
\begin{equation}
\begin{split}
&\mathbb{P}(||\boldsymbol{U}||_2 \leq \max_n(\epsilon_n)) \\
&=\mathbb{P}(||\boldsymbol{U}||_2 \leq \max_n[(\sigma_{re,n}+\sigma_{im,n})(\sqrt{a}+\sqrt{b}+\sqrt{2\ln(\frac{2N^2}{\delta})})] ) \\
&=\mathbb{P}(||\boldsymbol{U}||_2 \leq 1.4q(\sqrt{a}+\sqrt{b}+\sqrt{2\ln(\frac{2N^2}{\delta})}) ) \\
& \geq 1 - \sum_{n=1}^{N^2}[\delta_{n,re}+\delta_{n,im}] \\
& = 1 - \sum_{n=1}^{N^2}[\frac{\delta}{2N^2}+\frac{\delta}{2N^2}] = 1 - \sum_{n=1}^{N^2}\frac{\delta}{N^2} = 1 - \delta, 
\end{split}
\end{equation}
which implies the desired result.
\end{proof}

\subsection*{D. Putting everything together}

We now expand on the PAC-Bayes framework of \cite{mcallester1999some}.

\begin{customthm}{D.1}\label{pac_bayes}
(PAC-Bayes Theorem) Specifically let $f_{\bw}$ be any predictor (not necessarily a neural network) learned from the training data and parameterized by $\bw$. We assume a prior distribution $P$ over the parameters which should be a \emph{proper} Bayesian prior and cannot depend on the training data. We also assume a "posterior" $\mathcal{Q}$ over the predictors of the form $f_{\bw+\bu}$, where $\bu$ is a random variable which can have any distribution.  Then with probability at least $1-\delta$ we get
\begin{equation}\label{pac_bayes_equation}
\begin{split}
\mathbb{E}_{\bu} [ L_0(f_{\bw+\bu}) ] \leq & \mathbb{E}_{\bu}[\hat{L}_{0}(f_{\bw+\bu}) ] +\cO(\sqrt{\frac{2\text{KL}(\bw+\bu||P) +\text{ln} \frac{2 m }{\delta} }{ m-1  }}).
\end{split}
\end{equation}
\end{customthm}

Notice that the above gives a generalization result over a distribution of predictors. 

We now restate a usefull lemma which can be used to give a generalization result for a single predictor instance.
\begin{customlem}{4.2}\label{neysh_derandom}
\cite{neyshabur2017pac} Let $f_{\bw}(\bx):\mathcal{X} \Rightarrow \mathbb{R}^k$ be any predictor (not necessarily a neural network) with parameters $\bw$, and $P$ be any distribution on the parameters that is independent of the training data. Then with probability $\geq 1-\delta$ over the training set of size $m$, for any random perturbation $\bu$ s.t. $\mathbb{P}_{\bu}[\max_{\boldsymbol{x \in \mathcal{X}}} |f_{\bw+\bu }(\bx)-f_{\bw}(\bx)|_2 \leq \frac{\gamma}{4} ] \geq \frac{1}{2}$, we have 
\begin{equation}
L_0(f_{\bw}) \leq \hat{L}_{\gamma}(f_{\bw})+\cO(\sqrt{\frac{ \normalfont{\text{KL}}(\bw+\bu||P) +\text{ln} \frac{6 m }{\delta} }{ m-1  }}) 
\end{equation}
where $\gamma,\delta > 0$ are constants.
\end{customlem}

Contrary to Theorem \ref{pac_bayes}, Lemma \ref{neysh_derandom} links the empirical risk $\hat{L}_{\gamma}(f_{\bw})$ of the predictor to the true risk $L_0(f_{\bw})$, for a specific predictor and not a posterior distribution of predictors. We have also moved to using a margin $\gamma$ based loss. The perturbation $\bu$ quantifies how the true risk would be affected by choosing a bad predictor. The condition $\mathbb{P}_{\bu}[\max_{\boldsymbol{x \in \mathcal{X}}} |f_{\bw+\bu }(\bx)-f_{\bw}(\bx)|_2 \leq \frac{\gamma}{4} ] \geq \frac{1}{2}$ can be interpreted as choosing a posterior with \emph{small variance}, sufficiently concentrated around the current empirical estimate $\bw$, so that we can remove the randomness assumption with high confidence.

\begin{figure*}[t!]
\centering
\begin{subfigure}{.5\textwidth}
  \centering
  \includegraphics[scale = 0.46]{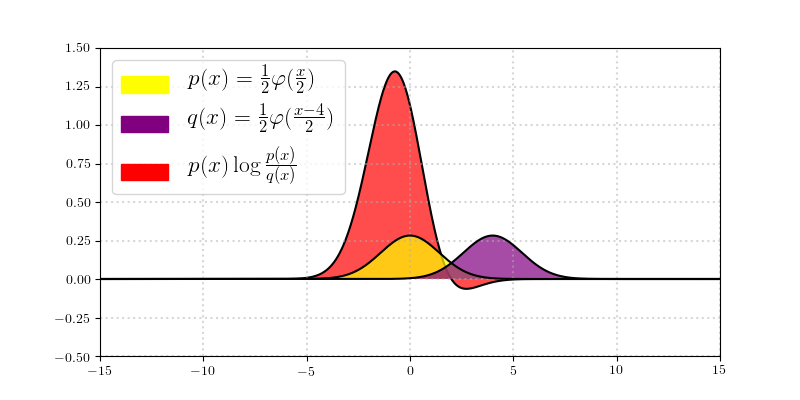}
  \caption{$\sigma = 2$ }
  \label{fig:Gaussians_1}
\end{subfigure}%
\begin{subfigure}{.5\textwidth}
  \centering
  \includegraphics[scale = 0.46]{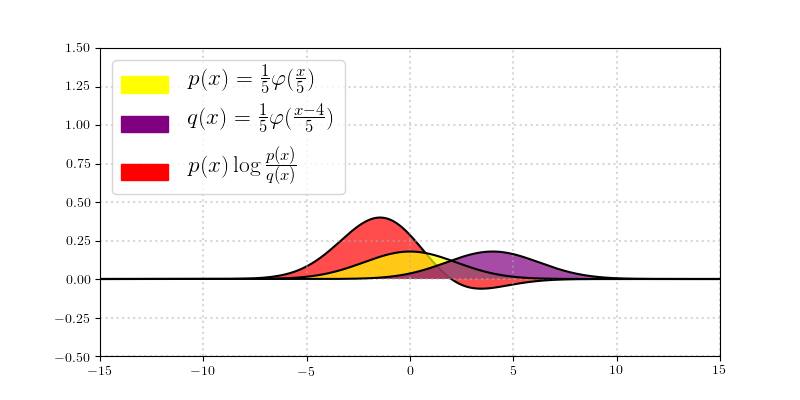}
  \caption{$\sigma = 5$}
  \label{fig:Gaussians_1}
\end{subfigure}
\caption{$D_{\text{KL}}(P||Q)$ \textbf{with} $p(x)=\frac{1}{\sqrt{2\pi\sigma^2}}e^{-\frac{x^2}{2\sigma^2}}$ \textbf{and} $q(x)=\frac{1}{\sqrt{2\pi\sigma^2}}e^{-\frac{(x-4)^2}{2\sigma^2}}$: By definition the KL divergence of the two distributions is the integral of the red curve. We see that as the variance increases the KL divergence between the two distributions decreases.}
\label{fig:Gaussians}
\end{figure*}

How small should we choose the the variance of $\bu$? The choice is complicated because the KL term in the bound is \emph{inversely proportional} to the variance of the perturbation (Figure \ref{fig:Gaussians}). Therefore we need to find the largest possible variance for which our stability condition holds.

Let $\beta = (\prod_{i=0}^d||\boldsymbol{W}_l||_2)^{1/d}$ and consider a network with the normalized weights $\tilde{\boldsymbol{W}}_l = \frac{\beta}{||\boldsymbol{W}_l||_2}\boldsymbol{w}_l$. Due to the homogeneity of the ReLu and Max-Pooling, we have that for feedforward neural networks with ReLu activations $f_{\tilde{\boldsymbol{w}}}=f_{\boldsymbol{w}}$ and so the (empirical and the expected) loss (including margin loss) is the same for $\tilde{\boldsymbol{w}} = \boldsymbol{w}$. We can also verify that $(\prod_{i=0}^d||\boldsymbol{w}_l||_2)=(\prod_{i=0}^d||\tilde{\boldsymbol{W}}_l||_2)$ and $\frac{||\boldsymbol{w}_l||_F}{||\boldsymbol{w}_l||_2}=\frac{||\tilde{\boldsymbol{W}}_l||_F}{||\tilde{\boldsymbol{W}}_l||_2}$, and so the excess error in the theorem statement is also invariant to this transformation. It is therefore sufficient to prove the theorem only for normalized weights $\tilde{\boldsymbol{w}}$, and hence we assume w.l.o.g. that the spectral norm is equal across layers, i.e. for any layer $i$, $||\boldsymbol{w}_l||_2$.

The prior cannot depend on the learned predictor $\boldsymbol{w}$ or it's norm, we will set $\sigma$ based on an approximation $\tilde{\beta}$ For each value of $\tilde{\beta}$ on a pre-determined grid, we will compute the PAC-Bayes bound, establishing the generalization guaranteee for all $\boldsymbol{w}$ for which $|\beta-\tilde{\beta}|\leq\frac{1}{d}\beta$, and ensuring that each relevant value of $\beta$ is covered by some $\tilde{\beta}$ on the grid. We will then take a union bound over all $\tilde{\beta}$ on the grid. In the previous we have considered a fixed $\tilde{\beta}$ and the $\boldsymbol{w}$ for which $|\beta-\tilde{\beta}|\leq\frac{1}{d}\beta$, and hence $\frac{1}{e}\beta^{d-1} \leq \tilde{\beta}^{d-1} \leq e \beta^{d-1}$.

Characterizing the condition  $\mathbb{P}_{\bu}[\max_{\boldsymbol{x \in \mathcal{X}}} |f_{\bw+\bu }(\bx)-f_{\bw}(\bx)|_2 \leq \frac{\gamma}{4} ] \geq \frac{1}{2}$ entails understanding the sensitivity of our classifier on random perturbations. 
To that end, we review here a usefull perturbation bound from \cite{neyshabur2017pac} on the output of a DNN:

\begin{customlem}{4.3}\label{sum_of_spectral_norms}
(Perturbation Bound). For any $B,d > 0$, let $f_w:\mathcal{X}_{B,n} \Rightarrow \mathbb{R}^k$ be a d-layer network with ReLU activations. Then, for any $\bw$, and $\bx \in \mathcal{X}_{B,n}$, and perturbation $\bu = \text{vec}(\{\bU_l \}^d_{l=1} )$ such that $||\bU_l ||_2 \leq \frac{1}{d}||\bW_l ||_2$, the change in the output of the network can be bounded as follows

\begin{equation}
  |f_{\bw+\bu }(\bx)-f_{\bw}(\bx)|_2 \leq e^2B \tilde{\beta}^{d-1} \sum_l ||\bU_l||_2
\end{equation}
where $e$, $B$ and $\tilde{\beta}^{d-1}$ are considered as constants after an appropriate normalization of the layer weights. 
\end{customlem}

We note that correctly estimating the spectral norm of the perturbation at each layer is critical to obtaining a tight bound. Specifically if we exploit the structure of the perturbation we can \emph{increase significantly} the variance of the added perturbation for which our stability condition holds.

We need to find the maximum variance for which $$\mathbb{P}_{\bu}[\max_{\boldsymbol{x \in \mathcal{X}}} |f_{\bw+\bu }(\bx)-f_{\bw}(\bx)|_2 \leq \frac{\gamma}{4} ] \geq \frac{1}{2}.$$ For this we will use Lemmas \ref{spectral_norm_for_convolutions} and \ref{fully_connected_spectral} which bound the spectral norm of the noise at each convolutional layer and sparse fully connected layer respectively. 

\begin{customlem}{4.5}
(Perturbation Bound). For any $B,d > 0$, let $f_w:\mathcal{X}_{B,n} \Rightarrow \mathbb{R}^k$ be a d-layer network with ReLU activations and we denote $\cC$ the set of convolutional layers and $\cF$ the set of fully connected layers. Then for any $\bw$, and $\bx \in \mathcal{X}_{B,n}$, and a perturbation for $\bu \sim \mathcal{N}(0,\sigma^2 \boldsymbol{I})$, for any $\gamma > 0$ with 

\begin{equation}
\sigma  = \frac{\gamma}{42 B \tilde{\beta}^{d-1} [\sum_{l\in \cC} K_l + \sum_{l\in \cF} J_l]}  
\end{equation}

we have

\begin{equation}
  \mathbb{P}_{\bu}[\max_{\boldsymbol{x \in \mathcal{X}}} |f_{\bw+\bu }(\bx)-f_{\bw}(\bx)|_2 \leq \frac{\gamma}{4} ] \geq \frac{1}{2}
\end{equation}
where $e$, $B$, $\tilde{\beta}^{d-1}$ are considered as constants after an appropriate normalization of the layer weights

\begin{equation}
K_l = q_l \{\sqrt{a_l}+\sqrt{b_l}+\sqrt{2\log(4N_i^2d)} \}
\end{equation}
and
\begin{equation}
J_l = q_l \{2\sqrt{s_l}+\sqrt{2\log(2d)} \}.
\end{equation}
\end{customlem}

\begin{proof}
We denote $\cC$ the set of convolutional layers, $\cF$ the set of fully connected layers and assume $|\cC|+|\cF|=d$ where $d$ is the total number of layers. We define events $||\bU_l||_2\leq 2\sqrt{s_l}+\sqrt{2\log(2d)}$ for the fully connected layers and $||\bU_l||_2\leq q_l\{\sqrt{a_l}+\sqrt{b_l}+\sqrt{2\log(4N_i^2d)}\}$ for the convolutional layers. We then assume that the probability for each of the $|\cF|$ and $|\cC|$ events is upper bounded by $\frac{1}{2d}$. We set $K_l = q_l \{\sqrt{a_l}+\sqrt{b_l}+\sqrt{2\log(4N_i^2d)} \}$ and $J_l = \{2\sqrt{s_l}+\sqrt{2\log(2d)} \}$ and take a union bound over the above events. After some calculations we obtain
\begin{equation}
\mathbb{P}(\sum_{i} ||\bU_l||_2 \leq \sigma[\sum_{l\in \cC} K_l + \sum_{l\in \cF} J_l]) \geq 1-(\sum_{l\in\cC}\frac{1}{2d}+\sum_{l\in\cF}\frac{1}{2d}) = 1 - \frac{1}{2} = \frac{1}{2}.
\end{equation}
We are then ready to apply our result directly to Lemma \ref{sum_of_spectral_norms}. We calculate that with probability $\geq \frac{1}{2}$
\begin{equation}
\begin{split}
|f_{\bw+\bu }(\bx)-f_{\bw}(\bx)|_2 &\leq e^2B \tilde{\beta}^{d-1} \sum_l ||\bU_l||_2 \\
&\leq \sigma e^2B \tilde{\beta}^{d-1} [\sum_{l\in \cC} K_l + \sum_{l\in \cF} J_l]. \\
\end{split}
\end{equation}

We have now found a bound on the perturbation at the final layer of the network as a function of $\sigma$ with probability $\geq \frac{1}{2}$. What remains is to find the specific value of $\sigma$ such that $|f_{\bw+\bu }(\bx)-f_{\bw}(\bx)|_2 \leq \frac{\gamma}{4}$. We calculate
\begin{equation}
\begin{split}
&|f_{\bw+\bu }(\bx)-f_{\bw}(\bx)|_2 \leq \frac{\gamma}{4} \\
&\Rightarrow \sigma e^2B \tilde{\beta}^{d-1} [\sum_{l\in \cC} K_l + \sum_{l\in \cF} J_l] \leq \frac{\gamma}{4}\\
&\Rightarrow \sigma \leq \frac{\gamma}{42 B \tilde{\beta}^{d-1} [\sum_{l\in \cC} K_l + \sum_{l\in \cF} J_l]}.\\
\end{split}
\end{equation}
\end{proof}

We can now calculate the KL term in Theorem 4.1. by noting that $\bw+\bu \sim \mathcal{N}(\bw,\sigma^2\boldsymbol{I})$, $P \sim \mathcal{N}(0,\sigma^2\boldsymbol{I})$, and that then $\text{KL}(\bw+\bu||P)\leq\frac{|\bw|^2}{2\sigma^2}$. We get that for any $\tilde{\beta}$, with probability $\geq1-\delta$ and for all $\bw$ such that, $|\beta-\tilde{\beta}|\leq\frac{1}{d}\beta$:
$$
L_0(f_{\bw}) \leq 
    \hat{L}_{\gamma}(f_{\bw}) 
    + \tilde{\cO}\left( \frac{B \, \Psi_f \, R_{\mathcal{W}}}{\gamma \sqrt{m} } \right),
$$
with $\|x\|_2 \leq B$ being a uniform bound on the input vectors,
$\Psi_f = q\sum_{l \in \cC} \sqrt{b_{l}} + \sum_{l \in \cF} \sqrt{s_l}, $
and
\begin{align}
    R_{\cW} := \prod_{l=1}^d||\bW_l||_2 \, \left( \sum_{l=1}^d  \frac{||\bW_l||_F^2}{||\bW_l||_2^2}\right)^{\hspace{-1mm}\frac{1}{2}}.
    \label{eq:spectral_complexity_ap}
\end{align}
Finally, we need to take a union bound over different choices of $\tilde{\beta}$. Let us see how many choices of $\tilde{\beta}$ we need to ensure we always have $\tilde{\beta}$ in the grid s.t. $|\beta-\tilde{\beta}|\leq\frac{1}{d}\beta$. We only need to consider values of $\beta$ in the range $(\frac{\gamma}{2B})^{1/d} \leq \beta \leq (\frac{\gamma\sqrt{m}}{2B})^{1/d}$. For $\beta$ outside this range the theorem statement holds trivially: Recall that the LHS of the theorem statement, $L_0(f_{\boldsymbol{w}})$ is always bounded by 1. If $\beta^d < \frac{\gamma}{2B}$, then for any $\boldsymbol{x}$, $|f_{\boldsymbol{w}}(\boldsymbol{x})| \leq \beta^dB \leq \gamma/2$ and therefore $L_{\gamma}=1$. Alternatively, if $\beta^d > \frac{\gamma\sqrt{m}}{2B}$, then the second term in equation \ref{pac_bayes_equation} is greater than one. Hence, we only need to consider values of $\beta$ in the range discussed above. Since we need $\tilde{\beta}$ to satisfy $|\beta-\tilde{\beta}|\leq\frac{1}{d}\beta \leq \frac{1}{d}(\frac{\gamma}{2B})^{1/d}$, the size of the cover we need to consider is bounded by $dm^{\frac{1}{2d}}$. Taking a union bound over the choices of $\tilde{\beta}$ in this cover and using the bound $L_0(f_{\bw}) \leq 
    \hat{L}_{\gamma}(f_{\bw}) 
    + \tilde{\cO}\left( \frac{B \, \Psi_f \, R_{\mathcal{W}}}{\gamma \sqrt{m} } \right),$ gives us the theorem statement. 

\section*{E. Additional experiments on the Bartlett Metric}
We include a number of additional experiments on the metric by \cite{bartlett2017spectrally}. The experimental setup is identical to the own used for the Neyshabur metric. We note that the conclusions we can draw are similar in both cases. They indicate a limitations of spectral complexity based generalization bounds in general. 

\begin{figure*}[h!]
\centering
\hspace{-2mm}
\begin{subfigure}{.7\textwidth}
  \centering
  \includegraphics[trim=16mm 0 15mm 10mm,clip,scale=0.55]{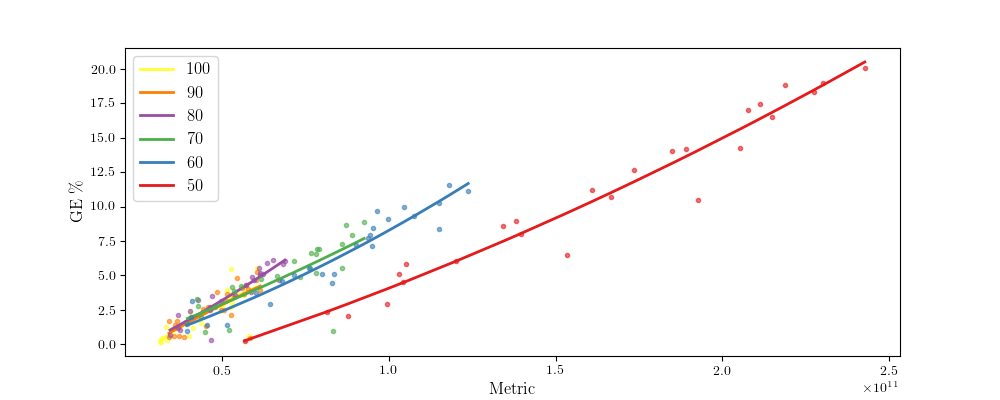}
  \caption{}
  \label{fig:conv_trans_var_metric_ap}
\end{subfigure}%
\begin{subfigure}{.3\textwidth}
  \centering
  \includegraphics[trim=1mm 0 0mm 10mm, clip,scale=0.55]{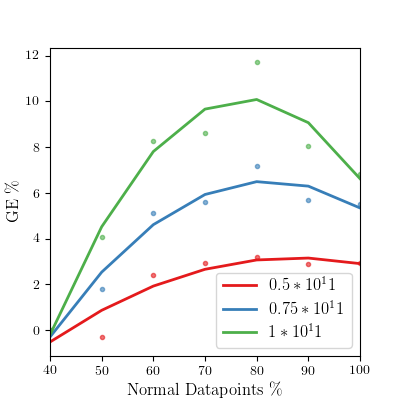}
  \caption{}
  \label{fig:conv_trans_fixed_metric_ap}
\end{subfigure}
\\
\vspace{1mm}
\hspace{-2mm}
\begin{subfigure}{.7\textwidth}
  \centering
  \includegraphics[trim=16mm 0 15mm 10mm,clip,scale=0.55]{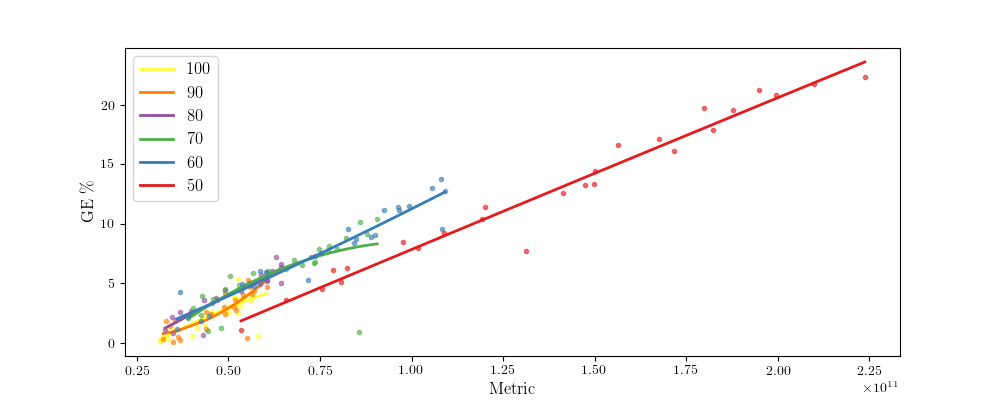}
  \caption{}
  \label{fig:conv_elastic_var_metric_ap}
\end{subfigure}%
\begin{subfigure}{.3\textwidth}
  \centering
  \includegraphics[trim=1mm 0mm 0mm 10mm, clip,scale=0.55]{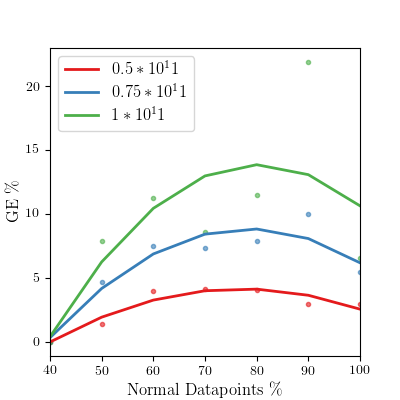}
  \caption{}
  \label{fig:conv_elastic_fixed_metric_ap}
\end{subfigure}
\caption{\textbf{Varying the percentage of translations (a-b) and elastic deformations (c-d)}: We split Training and Testing datasets of constant size into two parts---the first contains images that form a base space, whereas the rest of the dataset contains images that are augmentations of the base space. The percentage values indicate the percentage of the \emph{augmentations} over the total dataset. (a/c) We plot the GE vs spectral complexity. As we increase the number of translations/elastic deformations (equivalently decrease the percentage of the base space) the slopes of the GE curves decrease and we tend to have lower GE for the same spectral complexity metric values. (b/d) We plot the GE vs \% of augmentations for constant complexity values. The percentage of augmentations correlates empirically with the GE indicating that spectral complexity does not account for the architecture invariances.\label{fig:augmentations}}
\end{figure*}

\end{document}